\documentclass[lettersize,journal]{IEEEtran}
\usepackage{amsmath,amsfonts}
\usepackage{array}
\usepackage{textcomp}
\usepackage{stfloats}
\usepackage{url}
\usepackage{verbatim}
\usepackage{graphicx}
\usepackage{cite}
\usepackage{mathrsfs}
\usepackage{adjustbox}
\usepackage{multirow}
\usepackage{amssymb} 

\usepackage{color}
\usepackage[switch]{lineno}

\usepackage{mathrsfs}
\usepackage{amssymb}
\usepackage[figuresright]{rotating}
\usepackage[tableposition=top]{caption}
\usepackage{amsmath,epsfig,bbding}
\usepackage{epstopdf}
\usepackage{multicol}
\usepackage{booktabs}
\usepackage{multirow}
\usepackage{threeparttable}
\usepackage{color}
\DeclareCaptionFont{blue}{\color{blue}}
\usepackage{makecell}
\usepackage{amsmath}
\usepackage[linesnumbered,ruled,vlined]{algorithm2e}
\usepackage{algpseudocode}
\usepackage{arydshln}
\usepackage{makecell}
\usepackage{ulem}
\usepackage{bm}
\usepackage{xcolor}
\usepackage{subfigure}

\begin{document}

\title{CLIPVQA: Video Quality Assessment via CLIP}

\author{Fengchuang Xing,~
        Mingjie Li,~
         Yuan-Gen Wang,~\IEEEmembership{Senior Member,~IEEE,}
        Guopu Zhu,~\IEEEmembership{Senior Member,~IEEE},
        and~ Xiaochun Cao,~\IEEEmembership{Senior Member,~IEEE} 
\IEEEcompsocitemizethanks{

\IEEEcompsocthanksitem Fengchuang Xing, Mingjie Li, and Yuan-Gen Wang are with the School of Computer Science and Cyber Engineering, Guangzhou University, Guangzhou 510006, China (e-mail: xfchuang@e.gzhu.edu.cn; limingjie@gzhu.edu.cn; wangyg@gzhu.edu.cn). 
\IEEEcompsocthanksitem Guopu Zhu is with the School of Cyberspace Security, Harbin Institute of Technology, Harbin 150001, China (e-mail: guopu.zhu@hit.edu.cn).
\IEEEcompsocthanksitem Xiaochun Cao is with the School of Cyber Science and Technology, Shenzhen Campus of Sun Yat-sen University, Shenzhen 518107, China (e-mail: caoxiaochun@mail.sysu.edu.cn). 


}
}

\markboth{Journal of \LaTeX\ Class Files,~Vol.~14, No.~8, August~2021}%
{Shell \MakeLowercase{\textit{et al.}}: CLIPVQA}


\maketitle

\begin{abstract}
In learning vision-language representations from web-scale data, the contrastive language-image pre-training (CLIP) mechanism has demonstrated a remarkable performance in many vision tasks. However, its application to the widely studied video quality assessment (VQA) task is still an open issue. In this paper, we propose an efficient and effective CLIP-based Transformer method for the VQA problem (CLIPVQA). Specifically, we first design an effective video frame perception paradigm with the goal of extracting the rich spatiotemporal quality and content information among video frames. Then, the spatiotemporal quality features are adequately integrated together using a self-attention mechanism to yield video-level quality representation.
To utilize the quality language descriptions of videos for supervision, we develop a CLIP-based encoder for language embedding, which is then fully aggregated with the generated content information via a cross-attention module for producing video-language representation. Finally, the video-level quality and video-language representations are fused together for final video quality prediction, where a vectorized regression loss is employed for efficient end-to-end optimization.
Comprehensive experiments are conducted on eight in-the-wild video datasets with diverse resolutions to evaluate the performance of CLIPVQA.
The experimental results show that the proposed CLIPVQA achieves new state-of-the-art VQA performance and up to $37\%$ better generalizability than existing
benchmark VQA methods. A series of ablation studies are also performed to validate the effectiveness of each module in CLIPVQA.
\end{abstract}

\begin{IEEEkeywords}
Video Quality Assessment, In-the-wild Videos, Self-attention, Transformer, CLIP.
\end{IEEEkeywords}

\section{Introduction}\label{sec:introduction}
\IEEEPARstart{W}{ith} the advances in internet and video compression techniques, user-generated video content is growing exponentially on various social media platforms such as TikTok, YouTube, Meta, and Twitter. And also, the emergence of high-definition photography devices significantly increases the resolutions of videos created by professional or amateur users (e.g., $1080$P, $4$K, or even $8$K videos). The rapidly growing volume of videos and their continuously increasing resolutions in-the-wild has posed great challenges to the computer vision system, especially the video quality assessment (VQA). Thus, it is essential to develop efficient and effective VQA methods for these real-world videos, so as to better meet the demands of quality-of-experience (QoE) from users.

The ideal VQA methodology is subjective quality evaluation, which however is labor-intensive and
extremely time-consuming. As an alternative, the goal of objective VQA is to automatically and accurately predict the perceived quality of videos, thus attracting the interest of researchers.
According to the availability of the pristine reference videos, VQA can be classified into three categories: full-reference VQA, reduced reference VQA, and no-reference/blind VQA. The first two VQA paradigms heavily or partially rely on the referenced undistorted videos for generating quality predictions, which are impractical for scenarios where the pristine videos are inaccessible or even unavailable \cite{li2022blindly,jiang2022self}. Therefore, substantial efforts have been devoted to the no-reference VQA in recent years, which is also the focus of this paper\footnote{For simplicity, we directly use VQA to represent the no-reference/blind VQA in the following descriptions.}.
However, quantifying the perceptual quality of videos captured in-the-wild is a challenging task due to the absence of pristine references or shooting distortions.

Early VQA approaches are mainly based on handcrafted features to model particular types of distortions in videos, such as blockiness \cite{zhu2014no}, blur \cite{li2016spatiotemporal}, ringing \cite{rohil2020improved}, banding \cite{tu2020bband}, and noise \cite{norkin2018film}, for quality assessment.
However, these distortion-specific methods are unable to handle well the videos with diverse content and degradation that are naturally introduced during video acquisition.
Due to the prominent representation learning capability,
convolutional neural networks (CNNs) based methods have been demonstrated to be effective for the VQA task in recent years.
Instead of utilizing handcrafted features, CNNs-based VQA methods aim at extracting spatiotemporal features from videos by using deep convolutional neural networks \cite{kim2018deep, LSVQ, you2019deep, VSFA, MDVSFA,lu2022deep,shen2022end}. The extracted features are then regressed for predicting video quality.
The spatiotemporal features can be enriched for more effective VQA by integrating semantics-aware features \cite{tu2021rapique} and motion features \cite{li2022blindly}.
However, CNNs-based VQA methods have limited capacity to capture long-range dependencies, affecting their practical applications on high-resolution videos. 

Recently, Transformers \cite{Transformer} has demonstrated a remarkable self-attention mechanism for temporal modeling in many tasks, including language modeling \cite{dai2019transformer,cao}, image classification, and video processing \cite{timesformer}. To tackle the limitations of CNNs-based methods, Transformer-based approaches were proposed for VQA problem in-the-wild \cite{you2021long, xing2022starvqa, wu2022discovqa, li2022dcvqe, wu2022fast}.
A number of Transformer variants and attention strategies were introduced, including long-short-term convolutional Transformer \cite{you2021long}, divided space-time attention \cite{xing2022starvqa}, and fragment attention network \cite{wu2022fast}.
However, the above methods usually adopt pre-trained networks on classification datasets for the subsequent VQA prediction, which inevitably suffers from the problem of distributional shifts \cite{li2022blindly} between source (e.g., image classification) and target domains (in-the-wild VQA). This issue would result in sub-optimal generalizability.

Recent studies~\cite{radford2021learning, jia2021scaling} have shown the viability of the large-scale contrastive language-image pre-training (i.e., CLIP) method on computer vision applications. The key idea is to utilize web-scale image-text data, supervised by natural language, to learn visual or visual-text representations. After pre-training, the language information is applied to indicate these learned visual concepts, enabling the model to be transferred to downstream tasks, such as point cloud understanding~\cite{zhang2022pointclip}, dense prediction~\cite{rao2022denseclip}, image quality assessment~\cite{wang2023exploring,zhang2023blind}, and video recognition~\cite{zhou2022learning,ni2022expanding,ju2022prompting}. Motivated by this, in this paper, we propose to leverage video quality language as supervision to devise a novel video representation learning paradigm for VQA scenario to overcome the problem of distributional divergence between source and target domains.
Since this practice necessitates a considerable amount of video-language pre-training data and a large number of GPU resources (e.g., thousands of GPU days), directly training a video-language model is unfeasible.
Therefore, adapting the pre-trained image-language models to video domains is an alternative.
However, the application of the pre-trained image-language model to VQA is very limited in literature.
Two major challenges need to be overcome when transferring the pre-trained models from image to video domains: how to effectively i) extract the spatiotemporal information contained in videos and ii) learn the appropriate language representation for the final video quality estimation.

\begin{table}[tp]
\centering
\caption{\small Five-grade Quality Scale. }\label{quality scale}
\begin{tabular}{c:c:c}
\Xhline{1pt} %
Score  & Quality    & Impairment Description    \\
\hdashline
5   & Excellent  & Imperceptible    \\
4   & Good       & Perceptible, but not annoying  \\
3   & Fair       & Slightly annoying    \\
2   & Poor       & Annoying      \\
1   & Bad        & Very annoying   \\
\Xhline{1pt} %
\end{tabular}
\vspace{-2em}
\end{table}

\begin{figure*}[t]
\centering
\subfigure[Clear, well-lit, and good contrast. (MOS: 3.82)]{
\includegraphics[scale=0.4]{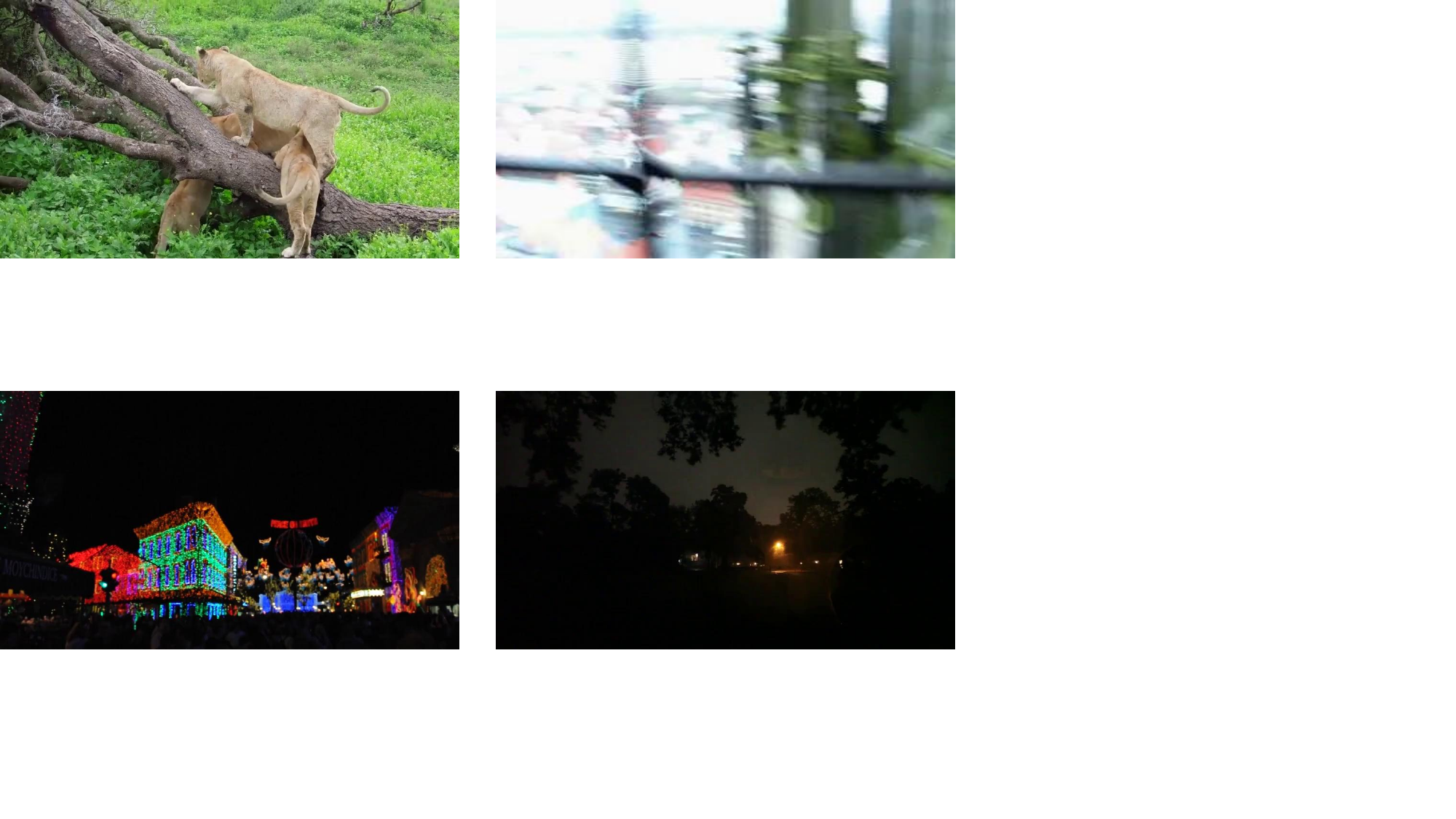}}
\subfigure[Not very clear and blurry. (MOS: 2.92)]{
\includegraphics[scale=0.4]{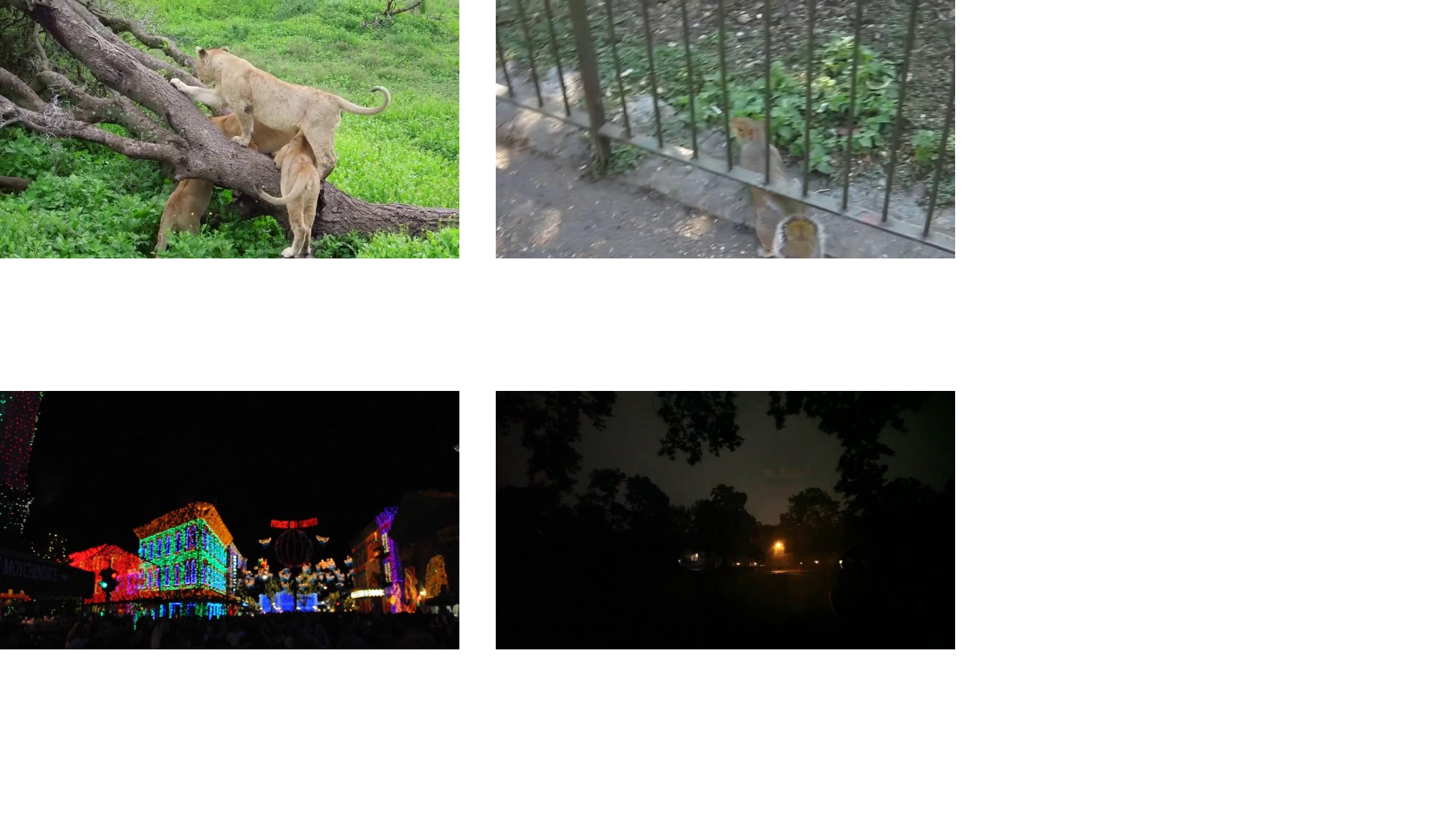}}
\subfigure[Vibrant, colorful, and vivid. (MOS: 3.18)]{
\includegraphics[scale=0.4]{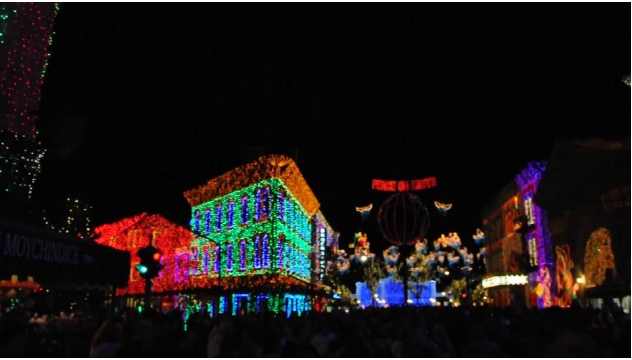}}
\subfigure[Relatively dark, dim, and murky. (MOS: 2.76)]{
\includegraphics[scale=0.4]{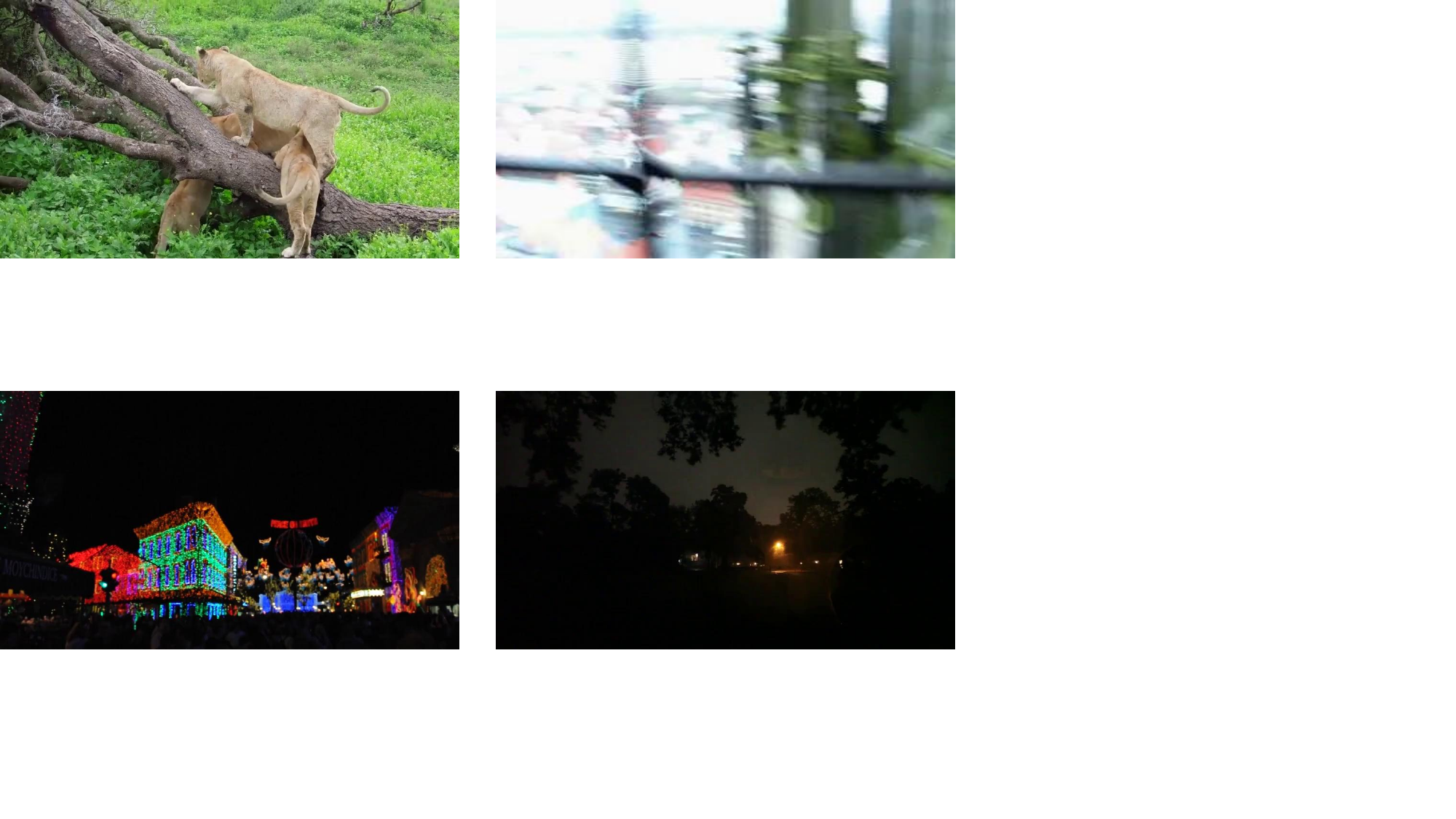}}
\vspace{-0.5em}
\caption{\small An illustration of the advantage of using natural language as supervision. These video frames are extracted from two natural VQA datasets, along with their corresponding quality text descriptions generated by a vision-language model BuboGPT\cite{zhao2023bubogpt}. With the help of these text descriptions strongly correlated to actual subjective assessment, a VQA model is more likely to make a better assessment that (a) and (c) are high-quality while (b) and (d) are low-quality frames.}
\label{LLaVA}
\vspace{-1em}
\end{figure*}

To address the first challenge, we propose a novel CLIP-based video frame perception Transformer (FPT).
It takes the sampled frames as input and aims to produce the frame tokens, pseudo Mean Opinion Score (MOS) tokens, and fusion tokens representations using a pre-trained image-language model. During the processing, the pseudo-MOS tokens are learned from the frame tokens, while the fusion tokens are derived from pseudo-MOS tokens. In a nutshell, these three representations are iteratively optimized from each other based on the attention-based schemes in order to realize that: i) The pseudo-MOS token of each frame not only contains the quality information of current frame, but also is able to communicate with other frames. ii) The fusion tokens carry both the intra-frame spatial and inter-frame temporal information of videos.
Then, the pseudo-MOS and fusion tokens are used together to yield the video-level quality representation by a spatiotemporal quality aggregation Transformer (SAT).

\begin{figure*}[tp]
\centering
\includegraphics[width=0.75\textwidth]{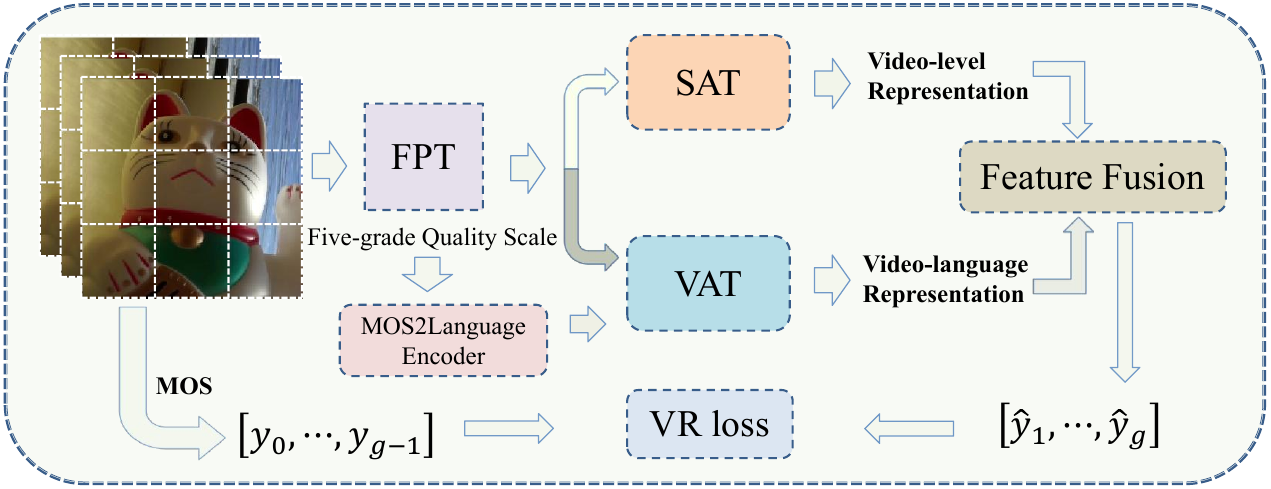}
\caption{\small An overview of the proposed CLIPVQA framework. It includes a frame perception Transformer (FPT), a spatiotemporal quality aggregation Transformer (SAT), a MOS2Language encoder, a video content and language aggregation Transformer (VAT), a fusion operation, and a vectorized regression (VR) loss for optimization.}\label{fig1}
\vspace{-1.5em}
\end{figure*}

To address the second challenge, we propose to take advantage of the natural language descriptions of video quality as supervision to learn representations for the VQA task. Our motivation is that with the help of quality text descriptions strongly correlated to actual subjective assessment, a VQA model is more likely to make a better quality evaluation for a video (See Fig.~\ref{LLaVA} for an illustration). Based on this, we first transform the MOS of videos into quality language descriptions using the ``Quality'' and ``Impairment Description'' texts as shown in Table~\ref{quality scale}. Then, we feed these language descriptions into a CLIP-based MOS2Language encoder to achieve the language representation.

After that, we design a video content and language aggregation Transformer (VAT) to make use of both the quality language embedding and the previous frame token representation to produce the video-language representation. Finally, the video-level and video-language representations are fused together to make the final video quality prediction. A vectorized regression loss function is utilized for the entire model optimization. We make the attempt to leverage natural language as supervision to tackle the VQA problem and our empirical studies verify its effectiveness.
Our contributions are summarized as follows:

\begin{itemize}
\item[$\bullet$] We propose an efficient and effective video frame perception paradigm for the VQA task to extract the spatiotemporal quality and content information from videos. 
It is mainly based on a CLIP-style mechanism for pre-training and can be seamlessly plugged into existing Transformer networks.

\item[$\bullet$] To deeply integrate the spatiotemporal features and the video quality information, we devise a spatiotemporal quality aggregation module for generating video-level quality representation.

\item[$\bullet$] We design a MOS2Language encoder to effectively embed the quality language descriptions of videos into language representation. With this language embedding as supervision, we develop a video content and language aggregator to yield video-language representation, which is fused with the video-level representation for final video quality prediction.

\item[$\bullet$] We conduct comprehensive experiments on diverse in-the-wild video datasets to evaluate the performance of the proposed framework. The results demonstrate that our proposed method achieves state-of-the-art VQA performance compared with three leading types of benchmark VQA methods.
\end{itemize}

The rest of this paper is organized as follows. First, we review the related works on VQA methods in Section~\ref{sec:relatedwork}. Then, we present the details of the proposed CLIP-based VQA method in Section~\ref{sec:proposed}. The experiments and analyses are conducted in Section~\ref{sec:exp}. Finally, we conclude this paper in Section~\ref{sec:conclusion}.

\section{Related Work}
\label{sec:relatedwork}

As this paper focuses on blind video quality assessment (VQA), in this section, we provide an overview of the relevant literature on blind VQA methods according to two sub-categories: classical VQA methods and deep VQA methods. Deep VQA methods can be further divided into CNNs-based and Transformer-based VQA methods. In addition, we also review some multi-modal VQA approaches and related applications. 

\subsection{Classical VQA Methods}
Early VQA methods were predominantly distortion-specific, focusing on banding \cite{tu2020bband}, noise \cite{norkin2018film}, blur \cite{marziliano2002no}, ringing \cite{feng2006measurement}, and blockiness \cite{wang2000blind}. However, these methods failed to take the overall perceptual quality of videos into consideration.
Afterwards, general-purpose VQA methods were developed to address the problems posed by diversified distortions. Based on scene statistic model \cite{ruderman1994statistics}, the natural scene statistics (NSS) methods \cite{sheikh2006image} were first applied to the domain of image quality assessment and then extended to natural bandpass space-time video statistics models \cite{li2016spatiotemporal,xue2014blind}.
Based on this paradigm, a number of methods were developed. VIIDEO method \cite{mittal2015completely} makes use of intrinsic statistical regularities to quantify distortions but does not involve any spatial features.
Li \textit{et al.} \cite{li2015no} proposed to leverage a model combining a 3D shearlet Transform to extract natural scene statistics.
Some approaches proposed to design separate spatial-temporal statistics \cite{yu2021predicting,chen2020perceptual}, where spatial features can be modified to capture the temporal effects within IQA models by employing simple or spatially displaced frame-differences \cite{yu2021predicting,lee2020video}. Korhonen \cite{TLVQM} proposed a two-level feature extraction mechanism to efficiently yield the distortion-relevant features. Tu \textit{et al.} \cite{VIDEVAL} introduced a fusion-based VQA model involving a feature ensemble and selection procedure.
Recently, Tu \textit{et al.} \cite{tu2021rapique} implemented a spatial and temporal bandpass statistics model combining quality-aware scene statistics features with semantics-aware features to further improve the VQA performance.
However, these traditional VQA methods mainly rely on low-level features derived from scene statistics models for video evaluation, which lacks the ability to well capture the structural, semantic and temporal information of videos, limiting their practical performance.

\begin{table}[tp]
\centering
\caption{\small Summary of Notations}
\label{tab:notation}
\begin{tabular}{c:l}
\Xhline{1pt} %
\textbf{Notation} & Description  \\ \hdashline
$h$   & The height of a cropped video frame\\   
$w$   & The width of a cropped video frame \\  
$s$   & The patch size \\  
$d$   & The token dimensionality \\  
$N$   & The number of sampled video frames \\  
$P$   & The number of patches in a cropped frame \\ 
$K$   & The total number of tokens in a sampled video \\  
$L$   & The number of CAT blocks \\  
$g$   & The number of gradations in the quality scale \\  
$r$   & The length of a text encoding \\   
$B$   & The number of CandLA blocks \\ \Xhline{1pt} %
\end{tabular}
\vspace{-2em}
\end{table}

\subsection{Deep VQA Methods}
\subsubsection{CNNs-based VQA Methods}
To handle the limitations of classical VQA algorithms, deep convolutional neural networks (CNNs) based methods were investigated and developed, which take full advantage of the powerful representation learning ability of CNNs for the VQA task \cite{goring2018deviq, korhonen2020blind}. Kim \textit{et al.} \cite{kim2018deep} proposed an aggregation network based on deep CNNs, named DeepVQA, to learn the maps of spatiotemporal visual sensitivity. Liu \textit{et al.} \cite{liu2018end} introduced a multi-task CNNs-based framework for VQA problem.
You and Korhonen \cite{you2019deep} utilized a 3D convolution network to extract regional spatiotemporal features from small cubic video clips, which are then fed into a long short-term memory network for predicting the perceived video quality. Li \textit{et al.} \cite{VSFA} constructed a VSFA model with a pre-trained image classification CNN and a gated recurrent unit, which is further enhanced to MDVSFA \cite{MDVSFA} by employing a mixed training strategy.
Ying \textit{et al.} \cite{LSVQ} suggested extracting
2D and 3D features from two branches, and then combining
them through a temporal regressor for VQA prediction.
Li \textit{et al.} \cite{li2022blindly} introduced a transfer learning-based model to extract motion features from a pre-trained large-scale action recognition model, which is then optimized by a mixed list-wise ranking loss function.
However, CNN-based VQA methods have a shortcoming in that they are unable to capture the long-range spatial dependencies within video frames, limiting their performance on high-resolution videos.

\subsubsection{Transformer-based VQA Methods}
Due to the success of Transformer on natural language processing and computer vision applications, Transformer-based VQA methods were studied.
You \cite{you2021long} introduced a perceptual hierarchical network with an attention module to extract quality features of video frames and fed them into a long-short-term convolutional transformer for video quality estimation.
Xing \textit{et al.} \cite{xing2022starvqa} designed a network by alternately concatenating a divided space-time attention and a vectorized regression loss for the VQA issue.
Wu \textit{et al.} \cite{wu2022discovqa} employed a Transformer-based model to capture the temporal relationships between video frames and their impacts on VQA.
Li \textit{et al.} \cite{li2022dcvqe} first split the whole video sequence into a number of clips followed by a Transformer to learn the clip-level quality, and then updated the frame-level quality for generating the final video-level quality.
Wu \textit{et al.} \cite{wu2022fast,wu2023neighbourhood} proposed a grid mini-patch sampling method and constructed a fragment attention network to learn video-quality-related representations. Shen\textit{et al.} \cite{10375568} presented a multiscale spatiotemporal pyramid attention (SPA) mechanism to extract multiscale features and models cross-scale dependencies for VQA task. Transformer-based models have been successfully applied to the VQA task and shown to perform better than CNNs-based VQA methods.
However, they suffer from the limitation of relying on classification datasets for pre-training network models.

Very recently, the contrastive language-image pretraining paradigm, i.e., CLIP, has demonstrated impressive generalization capacities on many computer vision tasks \cite{radford2021learning,jia2021scaling,zhou2024adaptive}.
Consequently, numerous CLIP-based methods were developed for downstream vision tasks. For instance, CoOp \cite{zhou2022learning} and CLIP-Adapter \cite{gao20233clip} adapted the CLIP-like vision-language models for image recognition.
CLIP-IQA \cite{wang2023exploring} and LIQE \cite{zhang2023blind} modified CLIP for image quality assessment (IQA). DenseCLIP \cite{zhou2021denseclip} used CLIP-based transfer learning for dense prediction.
VideoCLIP \cite{xu2021videoclip} extended the image-level pre-training manner to video domain by substituting the image-text data with video-text pairs for video understanding task.
Furthermore, XCLIP \cite{ni2022expanding} was proposed by directly adapting the pre-trained language-image models to video recognition.
EVL \cite{lin2022frozen} employed a lightweight Transformer decoder and learned a query token to dynamically collect frame-level spatial features from a CLIP-based image encoder for video recognition.
Although CLIP-based methods have achieved
very good performance for many vision tasks, their application to VQA is very limited.
In this paper, we propose an effective CLIP-based Transformer method to well address the blind VQA problem.  

\subsection{Multi-modal VQA}
It is noteworthy that recent studies \cite{min2020multimodal,min2016fixation,min2020study,cao2023attention} focus on the multi-modal VQA that explores the interplay between audio and visual signals for serving the overall quality perception. A number of related assessment models were developed to fuse the audio and video qualities, indicating a promising research direction for quality assessment (QA) community. 
Due to the success of deep learning and the significance of VQA problem, VQA techniques have been applied into a wide range of real-world QA scenarios, including screen content QA \cite{zeng2022screen,min2017unified}, enhancement QA \cite{min2019quality,chen2014quality,min2018objective,gu2015analysis}, and light field QA \cite{min2020metric,paudyal2019reduced}.

\section{Proposed Method}
\label{sec:proposed}

In this section, we introduce the technical details of the proposed method. We first briefly overview the whole CLIPVQA framework, and then describe each component in detail.
We utilize bold lowercase letters to denote (column) vectors and bold or special uppercase letters to denote matrices. The key notations used in this paper are summarized in Table~\ref{tab:notation}.

\subsection{Framework Overview}
\label{subsec:proposed:overview}
The proposed CLIPVQA method is a CLIP-based Transformer framework, as shown in Fig.~\ref{fig1}, which mainly consists of four components. Specifically, a frame perception Transformer (FPT) is first performed on the videos to produce the frame token, pseudo-MOS token, and fusion token representations, respectively (Subsec.~\ref{subsec:proposed:frame}). Then, both the pseudo-MOS and fusion token representations are passed into a spatiotemporal information aggregation Transformer (SAT) for generating the video-level quality representation (Subsec.~\ref{subsec:proposed:spatiotemporal}). After that, a MOS2Language encoder is introduced for the quality language encoding (Subsec.~\ref{subsec:proposed:encoder}). This is then combined with the frame token representation to produce the video-language representation via a video content and language aggregation Transformer (VAT) (Subsec.~\ref{subsec:proposed:video}).
Finally, both the video-level and video-language representations are fused to achieve the predicted probability vector, which is passed into a vectorized regression loss function for the framework optimization (Subsecs.~\ref{subsec:proposed:fusion} and~\ref{subsec:proposed:loss}).
Each of these modules will be explained below.

\begin{figure*}[tp]
\centering
\includegraphics[width=0.91\textwidth]{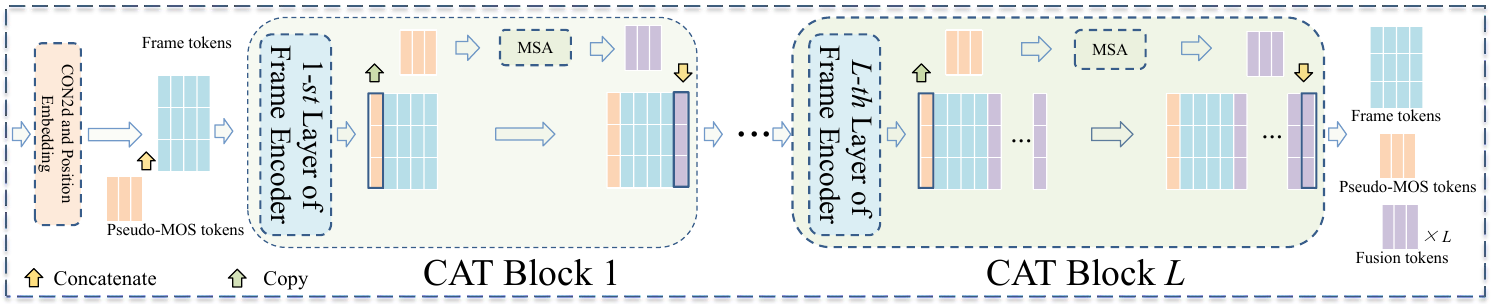}
\caption{\small An illustration of the frame perception Transformer (FPT), which consists of $L$ CAT blocks. The frame encoder is similar to the pre-trained image encoder in CLIP.}
\label{FTT}
\vspace{-1.5em}
\end{figure*}

\subsection{Video Frame Perception}
\label{subsec:proposed:frame}
The role of video frame perception is to fully capture the spatiotemporal information among video frames for subsequent video representation learning. 
To achieve this goal, we design a frame perception Transformer (FPT) which mainly consists of a certain number of CLIP-based aggressive Transformer (CAT) blocks. CAT block utilizes a CLIP-based frame encoder followed by a self-attention mechanism for representation learning.
Specifically, the FPT module first extracts the frame tokens by performing a convolutional network operation on selected video frames. Then it concatenates them with randomly generated pseudo-MOS tokens to serve as input for the CAT blocks. After undergoing the processes of the CAT blocks, FPT produces the frame and pseudo-MOS token representations and a kind of fusion token representation. We provide an illustration of FPT in Fig.~\ref{FTT}. The details of each part of FPT are introduced as follows.

\subsubsection{Frame Token Extraction}
We first randomly sample a preset number of frames from the video sequence with fixed intervals.
This equal-interval sampling scheme ensures frame diversity while reducing the redundancy within a video.
Then, each sampled frame is fed into a two-dimensional convolutional network, followed by embedding its positional information to produce the frame tokens.

To be specific, each sampled frame is randomly cropped with size $h \times w \times 3$, where $h$ and $w$ denote the height and width of the cropped frame, respectively, and $3$ represents the number of color channels. As shown in previous methods \cite{you2019deep, MDVSFA, LSVQ, tu2021rapique, li2022blindly}, by random cropping, a few dozens of training epochs are sufficient to cover most of the semantic information of the original frames. More importantly, the cropping operation can maintain the frame quality, which is naturally suitable for VQA tasks.

Subsequently, each cropped video frame is divided into non-overlapping patches with size $ s\times s$, and then we can obtain $ P=\lfloor h/s \rfloor \times \lfloor w/s\rfloor$ patches. Each patch is further flattened into a token with length $d = s\times s\times 3$. Suppose that we sample
$N$ video frames, $\mathbf{x}_{(n, p)}\in \mathbb{R}^d$ denotes the initial token for the $p$-th patch in the $n$-th selected frame, where $n\in[1,N]$ and $p\in[1,P]$. In this way, a sampled video contains a total number of $ K= N \times P$ tokens.
In order to capture the long-range dependence of spatial information through the self-attention mechanism, the spatial position information of each patch is also embedded to generate the frame tokens. Therefore, our frame token extraction is expressed as follows:
\begin{equation} \mathbf{e}_{(n,p)}=\mathbf{\Theta}\cdot \mathbf{x}_{(n,p)}+\mathbf{v}_{(n,p)},
\end{equation}
where $\mathbf{\Theta}\in \mathbb{R}^{d\times d}$ is a learnable matrix (i.e., the parameters of the two-dimensional convolutional network). $\mathbf{v}_{(n,p)}\in \mathbb{R}^d$ denotes the spatial position vector, which is generated using a sinusoid-based positional encoding \cite{BERT}. $\mathbf{e}_{(n,p)}\in \mathbb{R}^{d}$ is the generated frame token for the $p$-th patch in the $n$-th frame.

Finally, we randomly generate a pseudo-MOS token $\mathbf{\hat{e}}_{(n)} \in \mathbb{R}^d$ for each video frame and then concatenate it with the frame tokens from the associated frame, which is shown as below:
\begin{equation}\label{V2M}
\begin{split}
\mathbf{E}_{(n)}=\left[
 \begin{array}{c}
 \mathbf{\hat{e}}_{(n)}, \mathbf{e}_{(n,1)}, \dots, \mathbf{e}_{(n,P)}
 \end{array}
\right],
\end{split}
\end{equation}
where $\mathbf{E}_{(n)} \in \mathbb{R}^{d \times (P+1)}$ $(n \in [1, N])$. Subsequently, matrix $\mathbf{E}_{(n)}$ associated with the $n$-th video frame will be fed into the CAT blocks for further processing.

\subsubsection{CLIP-based Aggressive Transformer (CAT)}
The concatenation of frame and pseudo-MOS tokens will undergo the processes by a set number of CAT blocks to fully capture the underlying spatiotemporal information of the video sequence. Specifically, for the $1$-st CAT block, the matrix $\mathbf{E}_{(n)}$ $(n \in [1, N])$ is fed into the $1$-st layer of the frame-encoder. This frame-encoder is a pre-trained image encoder in CLIP, which mainly consists of multiple Transformer-based layers.
With the help of a pre-training scheme, the role of this frame encoder is to extract the intra-frame spatial information. Then, the output is rearranged into two matrices according to the pseudo-MOS and frame tokens, which can be expressed as follows (for the case of $l=1$):    
\begin{equation}\label{mos-token}
\mathcal{M}^{l}_{mos}=[
 \mathbf{\hat{e}}^{l}_{(1)},
 \mathbf{\hat{e}}^{l}_{(2)},P
  \dots,\mathbf{\hat{e}}^{l}_{(N)}],
\end{equation}
\begin{equation}\label{image-token1}
\mathcal{M}^{l}_{frame}=\left[
 [\mathbf{e}_{(1,1)}^{l},
  \dots,
 \mathbf{e}_{(1,P)}^{l}],
 \dots,
 [\mathbf{e}_{(N,1)}^{l},
  \dots,
 \mathbf{e}_{(N,P)}^{l}]
   \right],
\end{equation}
where $\mathcal{M}^{l}_{mos} \in \mathbb{R}^{d \times N}$ and $\mathcal{M}^{l}_{frame} \in \mathbb{R}^{d \times K}$ $(K=N\times P)$. The superscript $l$ denotes the numerical order of the current CAT block in FPT.

Next, $\mathcal{M}^{l}_{mos}$ is further processed by a single-layer multi-head self-attention (MSA) to achieve the fusion tokens, which is formulated as:
\begin{align}
\mathcal{M}_{fusion}^{l}&=\mathbf{MSA}(\mathbf{LN}(\mathcal{M}^{l}_{mos}))+\mathcal{M}^{l}_{mos} \\ \nonumber
&=[\mathbf{f}_{(1)}^{l}, \mathbf{f}_{(2)}^{l},
  \dots, \mathbf{f}_{(N-1)}^{l}, \mathbf{f}_{(N)}^{l}
   ],
\end{align}
where $\mathbf{MSA}(\cdot)$ denotes the multi-head self-attention operation and $\mathbf{LN}(\cdot)$ represents the LayerNorm operation. $\mathcal{M}_{fusion}^{l}\in\mathbb{R}^{d \times N}$ contains the fusion tokens of all video frames. This module enables the information exchange among multiple frames so as to capture the inter-frame temporal information. Then, for each frame, the fusion tokens are concatenated with the pseudo-MOS and frame tokens produced in the previous step to serve as the output of the current CAT block, which is expressed as:
\begin{equation}\label{V2M-2}
\mathcal{M}^{l}_{(n)}=\left[
 \mathbf{\hat{e}}_{(n)}^{l},
 \mathbf{e}_{(n,1)}^{l},
  \dots,
   \mathbf{e}_{(n,P)}^{l},
   \mathbf{f}_{(n)}^{1},
     \dots,
   \mathbf{f}_{(n)}^{l}
   \right],
\end{equation}
where $\mathcal{M}^{l}_{(n)} \in \mathbb{R}^{d \times (P+1+l)}$ $(n\in[1,N])$.
To acquire more temporal information from the videos, we design an \textbf{aggressive} strategy to maintain the fusion tokens generated by every CAT block. In other words, for each frame, a new fusion token will be generated and added to $\mathcal{M}^{l}_{(n)}$ when a new CAT block is executed.

To better capture the spatiotemporal features of the entire video, $L$ CAT blocks are sequentially performed to process the concatenation of frame, pseudo-MOS, and fusion tokens (i.e., $\mathcal{M}^{l}_{(n)}$). The number of the fusion tokens will gradually increase as $l$ raises. As this procedure progresses, the pseudo-MOS tokens gradually evolve into a state that reflects not only a single frame quality but also an encapsulation of the entire video quality.
This is achieved by enabling the exchange and integration of information among multiple pseudo-MOS tokens of frames, as well as the correlation between the pseudo-MOS and fusion tokens.

\subsubsection{Frame Token, Pseudo-MOS, and Fusion Token Representations}
After the deep processes of CAT blocks, the FPT module produces the frame token, pseudo-MOS, and fusion token representations, respectively, which are formulated as:
\begin{equation}\label{image-token}
\mathcal{M}^{L}_{frame}=\left[
 [\mathbf{e}_{(1,1)}^{L},
  \dots,
 \mathbf{e}_{(1,P)}^{L}],
   \dots,
  [\mathbf{e}_{(N,1)}^{L},
  \dots,
 \mathbf{e}_{(N,P)}^{L}]
   \right],
\end{equation}
\begin{equation}
\mathcal{M}^{L}_{mos}=[
 \mathbf{\hat{e}}^{L}_{(1)},
 \mathbf{\hat{e}}^{L}_{(2)},
  \dots,
   \mathbf{\hat{e}}^{L}_{(N)}],
\end{equation}
 \begin{equation}
\mathcal{M}_{fusion}^{L}=\left[
 [\mathbf{f}_{(1)}^{1},
  \dots,
 \mathbf{f}_{(1)}^{L}],
  \dots,
  [\mathbf{f}_{(N)}^{1},
   \dots,
   \mathbf{f}_{(N)}^{L}]
   \right] ,
\end{equation}
where $\mathcal{M}^{L}_{frame} \in \mathbb{R}^{d \times K}$ will be used for producing the video-language representation, $\mathcal{M}^{L}_{mos}\in\mathbb{R}^{d\times N}$ and $\mathcal{M}_{fusion}^{L} \in \mathbb{R}^{d\times(N\times L)}$ are then used for generating the subsequent video-level representation.

\subsection{Spatiotemporal Quality Aggregation}
\label{subsec:proposed:spatiotemporal}
After the frame perception stage, we achieve the pseudo-MOS and fusion token representations, which carry both rich spatiotemporal information and video quality-related information. Then, we design a spatiotemporal quality aggregation Transformer (SAT) to take full advantage of these representations for generating video-level quality representation. This SAT is mainly composed of a single-layer MSA and a multiple-layer perceptron (MLP).

Firstly, the pseudo-MOS token representation needs to be further transformed before aggregation. To more efficiently acquire the underlying temporal information, the location information also needs to be embedded into this pseudo-MOS representation. Specifically, this procedure is formulated as follows:
\begin{equation}
\mathcal{M}_{mos}=\mathcal{M}^{L}_{mos}+\mathbf{V}_{pos},
\end{equation}
\begin{equation}
\mathcal{M}_{mos}=\mathbf{LN}(\mathcal{M}_{mos} + \mathbf{MSA}(\mathbf{LN}(\mathcal{M}_{mos}))),
\end{equation}
where $\mathbf{V}_{pos} \in \mathbb{R}^{d \times N}$ denotes a position matrix generated using the sinusoid-based positional encoding \cite{BERT} and $\mathcal{M}_{mos} \in \mathbb{R}^{d\times N}$. Then, the fusion token representation is also processed by:
\begin{equation}
\mathcal{M}_{fusion}=\boldsymbol{\delta}\times \mathcal{M}_{fusion}^{L},
\end{equation}
where $\boldsymbol{\delta} \in \mathbb{R}^{d}$ represents a learnable weighted coefficient vector.

Finally, $\mathcal{M}_{fusion}$ and $\mathcal{M}_{mos}$ are aggregated together to produce the final representation, which is expressed as:
\begin{equation}
\mathbf{y}_{v}=\mathbf{Mean}(\mathcal{M}_{fusion} + \mathcal{M}_{mos}),
\end{equation}
where $\mathbf{y}_{v} \in \mathbb{R}^{r}$ is the video-level quality representation with length $r$, and $\mathbf{Mean(\cdot)}$ represents a mean operation.

\begin{figure*}[t]
\centering
\includegraphics[width=0.88\textwidth]{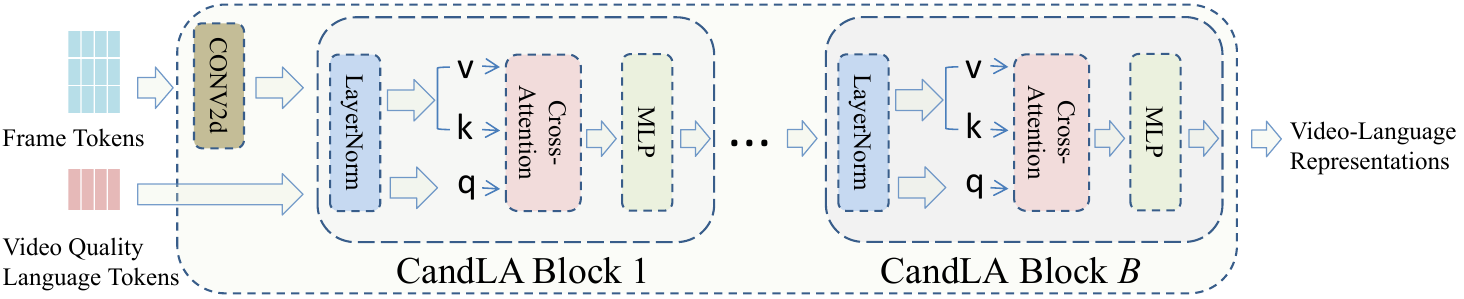}
\caption{\small An illustration of video content and language aggregation Transformer (VAT), which contains $B$ CandLA blocks.}\label{TFC}
\vspace{-1.5em}
\end{figure*}

\begin{table}[t]
\centering
\caption{\small Score and Long Language Description.}\label{text description-1}
\begin{tabular}{c:c}
\Xhline{1pt} %
Score  &  Long Language Description     \\
\hdashline
5   & Excellent and Imperceptible    \\
4   & Good and perceptible, but not annoying  \\
3   & Fair and slightly annoying    \\
2   & Poor and annoying      \\
1   & Bad and very annoying   \\
\Xhline{1pt} %
\end{tabular}
\vspace{-1.5em}
\end{table}

\begin{table}[t]
\centering
\caption{\small Score and Short Language Description. }\label{text description-2}
\begin{tabular}{c:c}
\Xhline{1pt} %
Score  &  Short Language Description     \\
\hdashline
5   & Excellent     \\
4   & Good  \\
3   & Fair    \\
2   & Poor      \\
1   & Bad    \\
\Xhline{1pt} %
\end{tabular}
\vspace{-2em}
\end{table}

\subsection{MOS2Language Encoder}
\label{subsec:proposed:encoder}
Inspired by the successes of recent CLIP-based methods like CLIP \cite{radford2021learning}, CLIP-Adapter \cite{gao20233clip}, CLIP-IQA \cite{wang2023exploring}, and LIQE \cite{zhang2023blind} on vision tasks, we propose to use natural language as supervision to learn precise video representation for VQA scenario. The motivation is based on that a VQA model may make a better quality assessment for a video with the assistance of its quality text descriptions as shown Fig.~\ref{LLaVA}. 
To realize this goal, we design a CLIP-based encoder (called MOS2Language encoder) for quality text descriptions so as to provide language supervision for the subsequent representation learning.

Firstly, we need a five-grade quality scale as shown in Table \ref{quality scale}, which is essential for conducting subjective VQA process \cite{bt1788methodology}. To verify which part of the quality descriptions is more important, we divide the quality scale into two subtables according to the ``quality'' and ``impairment descriptions'', which are shown in Table \ref{text description-1} and Table \ref{text description-2}, respectively.
Then, we feed the language descriptions into the CLIP-based text
encoder for language encoding, which is expressed as:
\begin{equation}
\mathcal{Y}_{t}=\mathbf{Text\underline{~}Encoder}(\mathcal{T}),
\end{equation}
where $\mathcal{T}$ stands for the video quality text descriptions as shown in the second column of Tables~\ref{text description-1} and \ref{text description-2}. $\mathcal{Y}_{t}\in \mathbb{R}^{g \times r}$ is the matrix of the final encoded text, where $g$ represents the number of gradations in the quality scale and $r$ denotes the length of the encoding. 
The $\mathbf{Text\underline{~}Encoder}$ is mainly composed of several MSAs and will be pre-trained, which is similar to the text encoder in CLIP \cite{radford2021learning}. More details can be found in \cite{radford2021learning}. Note that the long language description (Table~\ref{text description-1}) and the short language case (Table~\ref{text description-2}) will be investigated separately in our experiments, which can be seen in Subsection \ref{subsec:exp:ablation}.

\subsection{Video Content and Language Aggregation}
\label{subsec:proposed:video}
By performing the frame perception Transformer (FPT), we obtain the frame token representation ($\mathcal{M}^{L}_{frame}$) that carries rich video content information.
In this stage, we introduce a video content and language aggregation Transformer (VAT) to make use of this content information for producing the video-language representation, utilizing the quality language encoding as supervision.
Specifically, VAT mainly consists of $B$ content and language aggregation (CandLA) blocks, which are illustrated in Fig. \ref{TFC}.
Each CandLA block contains a LayerNorm operation (LN) and a multi-head cross-attention (MCA), followed by a multiple-layer perceptron (MLP).

To reduce redundant information in the video frames and adapt to the input requirement of a CLIP-based network, $\mathcal{M}^{L}_{frame}$ is first processed using a two-dimensional convolutional network for dimensionality reduction:
\begin{equation}
\mathcal{Y}_{c}=\mathbf{CONV2d}(\mathcal{M}^{L}_{frame}),
\end{equation}
where $\mathcal{Y}_{c} \in \mathbb{R}^{g \times r}$ is the video content representation. Then $\mathcal{Y}_{c}$ and the quality language encoding $\mathcal{Y}_{t}$ are fed into the CandLA blocks for aggregation. For each CandLA block, the process can be formulated as:
\begin{equation}
\mathcal{\hat Y}_{v}=\mathcal{Y}_{t}+\mathbf{MCA}(\mathbf{LN}_q(\mathcal{Y}_{t}),\mathbf{LN}_k(\mathcal{Y}_{c}), \mathbf{LN}_v(\mathcal{Y}_{c})),
\end{equation}
\begin{equation}
\mathcal{\tilde Y}_{v}=\mathcal{\hat Y}_{v}+\mathbf{MLP}(\mathbf{LN}(\mathcal{\hat Y}_{v})),
\end{equation}
where $\mathcal{\hat Y}_{v}\in\mathbb{R}^{g \times r}$ denotes the output of $\mathbf{MCA}$. $\mathbf{LN}_q(\cdot)$, $\mathbf{LN}_k(\cdot)$, and $ \mathbf{LN}_v(\cdot)$ yield the
queries, keys, and values, respectively. $\mathcal{\tilde Y}_{v} \in \mathbb{R}^{g \times r}$ is the output of the CandLA block.
To fully integrate the language supervision and the content information, $B$ CandLA blocks are sequentially performed to produce a high-quality video-language representation.

\subsection{Feature Fusion}
\label{subsec:proposed:fusion}
To generate the video quality prediction, the video-level representation $\mathbf{y}_{v}\in\mathbb{R}^{r}$ produced by SAT and the video-language representation $\mathcal{\tilde Y}_{v} \in \mathbb{R}^{g \times r}$ obtained by VAT are fused via an operation of element-wise multiplication, which is expressed as:
\begin{equation}
\mathbf{\hat{y}}=\mathbf{Softmax}(\mathcal{\mathcal{\tilde Y}}_{v}\times \mathbf{y}_{v}),
\end{equation}
where $\mathbf{\hat{y}} \in \mathbb{R}^{g}$ denotes the predicted probability vector and $\mathbf{Softmax}(\cdot)$ stands for a Softmax activation function.
We hope this probability vector can well match the human assessor's quality rating distributions for a video.

\subsection{Vectorized Regression Loss}
\label{subsec:proposed:loss}
An objective VQA model is expected to produce quality predictions of videos that are highly consistent with subjective ratings.
Following the method in \cite{xing2022starvqa}, we employ a vectorized regression (VR) loss function to optimize our CLIPVQA model for higher assessment precision.
The key point is to transform the ground truth quality score into a vector, which is then integrated with the predicted probability vector to generate the final loss function.

Concretely, we first linearly scale the mean opinion score (MOS) value (i.e., the ground truth score) of each video within a given dataset into a unified interval $[T, U]$, where $T$ and $U$ represent the lowest and highest scaled MOS values in the dataset, respectively. In our experiments, $T$ and $U$ are respectively set to be $1$ and $5$. This practice can unify the MOS scales of different datasets.
Then, based on this interval $[T, U]$, we generate $g$ reference ratings and the $i$-th reference rating is calculated as
\begin{equation}
\label{eq:anchor}
b_i=T+\frac{U-T}{g-1} \times i,
\end{equation}
where $i \in \{0,1,...,g-1\}$. 
Note that $g$ is also the number of gradations in the quality description. 
These reference ratings are used to help encode the MOS value, which is inspired by the subjective rating process.
The reference ratings closer to MOS value are more likely to be selected by the human assessors.
Suppose that a scaled MOS of a video is denoted as $c$, a real number score. Then, it is encoded into a probability vector by
\begin{equation}
\label{mos}
\bf{y} = \mathbf{Softmax}(\bf{z}),
\end{equation}
where $\textbf{z} = [z_0,\dots,z_{g-1}] \in \mathbb{R}^{g}$ with $ z_i= -(c-b_i)^2$. $\mathbf{y} \in \mathbb{R}^{g}$ is the MOS encoding, among which the ratings closer to MOS value $c$ have higher probabilities.
This MOS encoding well reflects the actual ratings given by the assessors, thereby can naturally be used as pseudo labels for optimizing the framework.
Finally, the VR loss function is calculated between the
encoded MOS vector $\mathbf{y}$ and the predicted probability vector $\mathbf{\hat{y}}$, which is expressed as:
\begin{equation}\label{loss}
\mathcal{L}_{\textrm{VR}}=1-\frac{\langle\mathbf{y}\cdot\mathbf{\hat{y}}\rangle}{||\mathbf{y}||\cdot||\mathbf{\hat{y}}||},
\end{equation}
where the $\langle \cdot \rangle$ denotes the inner product operation and $||\cdot||$ is the $L_2$-norm operation.

In the inference stage, the model needs to output a quality score of a video for users. The rating with the highest probability in the predicted probability vector $\mathbf{\hat{y}}$ cannot be directly used as quality score, since the final quality score is a real number and also related to other ratings.
Thus, we employ a support vector regressor (SVR) to transform the $\mathbf{\hat{y}}$ into a real-valued quality score $\hat{c}$ which is used as final output. To accomplish this transformation, the samples ($\mathbf{y}$,$c$) need to be used for the SVR training. Note that it is possible to implement a reverse transformation module as a part of the overall model instead of using a separate SVR. For example, we can design a simple fully-connected network behind the feature fusion module to perform the function of SVR. According to our empirical studies, these two practices yield almost the same effects. However, the multiple linear layers inevitably increase the parameter volume and training costs, compared to offline and fast SVR.

\begin{table*}[ht]
   \centering
  \caption{\small Summary of the Benchmark VQA Datasets.}\label{dataset}
  \resizebox{1.9\columnwidth}{!}{
  \small
  \begin{tabular}{cllllllllll}
   \Xhline{1pt} %
   DATABASE &  YEAR & CONT & TOTAL & RESOLUTION   & FR & LEN  & FORMAT  & MOS Range & SUBJ   &ENV \\
   \midrule %
    CVD2014 \cite{CVD2014}   & 2014   & 5    &  234  & 720p,480p  & 9-30  & 10-25 & AVI   & [-6.50,93.38]  & 210         & In-lab \\ \hdashline   
    KoNViD-1k \cite{hosu2017konstanz}  & 2017   & 1,200 &  1,200 & 540p       & 24-30 & 8     & MP4   & [1.22,4.64]  & 642     & Crowd \\   \hdashline 
    LIVE-Qualcomm \cite{LIVE-Q} & 2018   & 54   &  208 & 1080p  & 30 & 15 & YUV & [16.5621,73.6428] & 39   & In-lab \\   \hdashline 
    LIVE-VQC \cite{LIVE-VQC}   & 2018   & 585   &  585 & 240p-1080p  & 19-30 & 10 & MP4 & [6.2237,94.2865] & 4776   & Crowd \\   \hdashline 
    YouTube-UGC \cite{UGC-2019}   & 2019   & 1,380   &  1,380 & 360p-4k  & 15-60 & 20 & MP4 & [1.242,4.698] & $>$8K  & Crowd \\  \hdashline  
    LSVQ \cite{LSVQ}   & 2020   & 39,075   &  39,075 & 99p-4k  & 25/50 & 5-12 & MP4 & [2.4483,91.4194] & 5.5M   & Crowd \\  \hdashline  
    LSVQ-1080p \cite{LSVQ}   & 2020   & 3,573   & 3,573 & 960p-4k  & 25/50 & 5-12 & MP4 & [17.7222,91.4194] & --   & Crowd \\  \hdashline  
    KonViD-150k \cite{gotz2021konvid} & 2021   & 153,841 &  153,841 & 540p  & 24-30 & 5     & MP4   & [1.0,5.0]  & --     & Crowd   \\    
   \Xhline{1pt}
  \end{tabular}
  }
 \begin{tablenotes}
 \item[1] \textit{CONT=Total number of unique contents}, \textit{TOTAL=Total number of test sequences, including reference and distorted videos}. \textit{RESOLUTION=Video resolution}, \textit{FR=Frame rate}, \textit{LEN=Video duration/length (in seconds)}, \textit{FORMAT=Video container}. \textit{MOS Range=Range of mean opinion score}, \textit{SUBJ=Total number of subjects in the study}. \textit{ENV=Subjective testing environment}. \textit{In-lab=Study is conducted in a laboratory}. \textit{Crowd=Study is conducted by crowdsourcing}.
 \end{tablenotes}
 \vspace{-0.5em}
 \end{table*}

\begin{table*}[htp]
\centering
\caption{\small Parameter Settings and Complexities of the CLIPVQA Variants.}\label{Implementation details}
\resizebox{1.9\columnwidth}{!}{
\begin{tabular}{l:lcccccc}
\Xhline{1pt}
    CLIP                  & Methods      &Frames ($N$) $\times$ Sample Rate   &  Size of Patch ($s$)      &Training Views      & Testing Views     & FLOPs    & Parameters   \\  \hline
\multirow{4}{*}{CLIP-B-16}
                        & CLIPVQA-16-8   & 8  \;$\times$ 4                    & 16 $\times$ 16              & 1 $\times$ 1    & 1 $\times$ 3      & 145G     & 188M       \\
                        & CLIPVQA-16-16  & 16 $\times$ 4                    & 16 $\times$ 16              & 1 $\times$ 1    & 1 $\times$ 3      & 287G     & 190M         \\
                        & CLIPVQA-16-32  & 32 $\times$ 4                    & 16 $\times$ 16              & 1 $\times$ 1    & 1 $\times$ 3      & 574G     & 194M           \\
                        & CLIPVQA-16-64  & 64 $\times$ 2                    & 16 $\times$ 16              & 1 $\times$ 1    & 1 $\times$ 3      & 1148G    & 201M           \\  \hdashline
\multirow{2}{*}{CLIP-L-14}
                        & CLIPVQA-14-8   & 8  \; $\times$ 4                  & 14 $\times$ 14              & 1 $\times$ 1    & 1 $\times$ 3      & 658G     & 553M     \\
                        & CLIPVQA-14-16  & 16 $\times$ 4                    & 14 $\times$ 14              & 1 $\times$ 1    & 1 $\times$ 3      & 1316G    & 555M     \\
\Xhline{1pt}
\end{tabular}
}
\begin{tablenotes}
 \item[1] \textit{Sample rate denotes the time interval of the sampling from the original video, and the training views and testing views represent how many times a frame is cropped from an original video frame in training and testing processes, respectively. A CLIPVQA variant is directly denoted by CLIPVQA-$s$-$N$.}
 \end{tablenotes}
 \vspace{-1.5em}
\end{table*}

\subsection{Comparison to CLIP-based IQA Methods}
CLIP-IQA \cite{wang2023exploring} and LIQE \cite{zhang2023blind} are two recent CLIP-based methods for image quality assessment (IQA) problems. However, CLIPVQA greatly differs from these two approaches by the fact that the way of using CLIP is different. This also leads to their architectures being distinct. 
Specifically, both CLIP-IQA and LIQE use CLIP in an \textit{explicit} way similar to the original CLIP, where an image encoder and a text encoder are responsible for extracting visual and textual information, respectively, followed by a fusion operation for final prediction. Instead, our method utilizes CLIP in an \textit{implicit} way where a pre-trained frame encoder integrated with a self-attention mechanism (i.e., CAT block) is designed for video frame perception and a MOS2Language encoder is developed for language embedding. This is then followed by two carefully designed aggregation modules to better serve the fusion action for the final prediction. Another disparateness is that the studied tasks are different. IQA only focuses on the spatial information of still images while VQA is concerned with the spatio-temporal information of motion videos. This results in different learning strategies and loss functions. For example, CLIP-IQA uses a prompt pairing strategy for text encoding while LIQE adopts a multitask learning scheme for the loss function. By contrast, our CLIPVQA employs a vectorized regression loss for efficient end-to-end optimization.    

\section{\textbf{Experimental Results and Analyses}}
\label{sec:exp}

In this section, we conduct comprehensive experiments to compare our CLIPVQA method with the \textit{state-of-the-art} VQA methods on the benchmark datasets.
We first outline the experimental settings (Subsec.~\ref{subsec:exp:setup}). Then we evaluate the performance of CLIPVQA through experiments on individual datasets and in cross-dataset setting, respectively (Subsecs.~\ref{subsec:exp:ind} and ~\ref{subsec:exp:cross}). We also evaluate our method by experiments in a fine-tuning manner (Subsec.~\ref{subsec:exp:fine}) and provide qualitative analyses for the predictions of CLIPVQA (Subsec.~\ref{subsec:exp:mos}). Finally, we perform ablation studies to verify the effectiveness of each component of CLIPVQA (Subsec.~\ref{subsec:exp:ablation}), followed by a computational cost analysis for the entire CLIPVQA model (Subsec.~\ref{subsec:exp:comp}).

\subsection{Experimental Setup}
\label{subsec:exp:setup}

\textbf{Datasets Description} 
The experiments are conducted on eight natural video benchmark datasets, including CVD2014 \cite{CVD2014}, KoNViD-1k \cite{hosu2017konstanz}, LIVE-Qualcomm \cite{LIVE-Q}, LIVE-VQC \cite{LIVE-VQC}, YouTube-UGC \cite{UGC-2019}, LSVQ \cite{LSVQ}, LSVQ-1080p \cite{LSVQ}, and KonViD-150k \cite{gotz2021konvid}. Their details are summarized in Table \ref{dataset}.
It can be seen that these datasets are heterogeneous in terms of content, resolution, frame rate, time duration, and format. Moreover, the LSVQ-1080p, i.e., a subset of LSVQ, is constructed with $3,573$ videos, where over 93\% of them have a resolution higher than $1080$p.
These datasets contain a total of about $200$k video samples, covering the most common resolutions, distortions, and content types. The experiments on these datasets can well evaluate the performance of VQA models.

\textbf{Comparison Methods} 
Our proposed CLIPVQA method is compared with a wide range of existing state-of-the-art VQA methods, including classical methods such as TLVQM \cite{TLVQM}, VIDEVAL \cite{VIDEVAL}, RAPIQUE \cite{tu2021rapique}, and BRISQUE \cite{mittal2012no}, CNNs-based methods like VSFA \cite{VSFA}, MDTVSFA \cite{MDVSFA}, PVQ \cite{LSVQ}, MLSP-FF \cite{gotz2021konvid}, GST-VQA \cite{chen2021learning}, CoINVQ \cite{wang2021rich}, CNN+TLVQM \cite{korhonen2020blind}, and BVQA-2022 \cite{li2022blindly}, and Transformer-based methods such as StarVQA \cite{xing2022starvqa}, DisCoVQA \cite{wu2022discovqa}, DCVQE \cite{li2022dcvqe}, and FAST-VQA \cite{wu2022fast}. 
We collect their reported experimental results for performance comparison.
For the baselines whose source codes are not publicly available online or cannot be obtained from authors, we do our utmost to implement them according to the original papers and conduct experiments for them.

\textbf{Implementation and Parameter Details} 
Our network is built on the Pytorch framework and trained with four Tesla 3090 GPUs.
The parameters of the comparison methods are set to the default values as suggested in their papers unless otherwise specified.
For our CLIPVQA, by default, the number of sampled frames $N=32$, the height ($h$) and width ($w$ ) of a cropped frame are both set to be $224$. The patch size $s$ is $16$. The numbers of CAT and CandLA blocks are set to be $12$ and $2$, respectively. 
Our method uses the two pre-trained models, CLIP-B-16 and CLIP-L-14 for training, where their layer numbers are set to be $12$ and $24$, respectively. 
Specifically, the parameters of the frame encoder in CAT blocks and the MOS2Language encoder are initialized by the pre-trained weights of the image encoder and the text encoder in CLIP, respectively. The parameters of the other components are randomly initialized.
During the end-to-end training, the parameters of MOS2Language encoder are frozen while the frame encoder is fine-tuned according to the back-propagation mechanism. Except for these two parts, the other parameters are trained from scratch.
This training strategy already enables the proposed framework to achieve very good performance on all datasets.
The learning rate is initialized at $0.005$, with a momentum of $0.9$ and a decay rate of $0.1$ for every $10$ epochs. The datasets are randomly split into a training set and a testing set by a ratio of $8:2$.
The other parameter settings of CLIPVQA are provided in Table \ref{Implementation details}.
For more detailed configuration information, please refer to our source code at \emph{https://github.com/GZHU-DVL/CLIPVQA}.

\textbf{Performance Metrics} 
To evaluate the VQA performance of the comparison methods, we adopt two widely used metrics: the Spearman's Rank-order Correlation Coefficient (SROCC) and the Pearson's Linear Correlation Coefficient (PLCC).

\begin{table*}[htp]
\centering
\caption{\small VQA Performance of All Comparison Methods on Individual Datasets.} \label{individual train}
\resizebox{1.9\columnwidth}{!}{
\small
\begin{tabular}{l:lcc|cc|cc|cc|cc}
\Xhline{1pt} %
\multicolumn{2}{c}{Type/ } &\multicolumn{8}{c}{Intra-dataset Test sets}        \\
\hdashline%
\multicolumn{2}{c}{Testing sets/} &\multicolumn{2}{c|}{LSVQ} &\multicolumn{2}{c|}{LSVQ-1080p} &\multicolumn{2}{c|}{KoNViD-1k} &\multicolumn{2}{c|}{LIVE-VQC} &\multicolumn{2}{c}{YouTube-UGC}  \\ \hline
   Groups & Methods                    &  SROCC & PLCC         &  SROCC & PLCC          &  SROCC & PLCC      &  SROCC & PLCC      &  SROCC & PLCC              \\    \hline
\multirow{3}{*}{\makecell[l]{Existing \\ Classical \\ Methods}}
                        & BRISQUE \cite{mittal2012no}      & 0.569 & 0.576         & 0.497 & 0.531          & 0.646 & 0.647      & 0.524 & 0.536      & NA & NA                     \\
                        & TLVQM \cite{TLVQM}    & 0.772 & 0.774         & 0.589 & 0.616          & 0.732 & 0.724      & 0.670 & 0.691      & 0.669 & 0.659                 \\
                        & VIDEVAL \cite{VIDEVAL}      & 0.794 & 0.783         & 0.545 & 0.554          & 0.751 & 0.741      & 0.630 & 0.640      & 0.778 & 0.773                    \\  \hdashline
\multirow{4}{*}{\makecell[l]{Existing \\ CNNs-based \\ Methods}}
                        & VSFA \cite{VSFA}        & 0.801 & 0.796         & 0.675 & 0.704          & 0.784 & 0.794      & 0.734 & 0.772      & 0.724 & 0.743                   \\
                        & PVQ-wo \cite{LSVQ}      & 0.814 & 0.816         & 0.686 & 0.708          & 0.781 & 0.781      & 0.747 & 0.776      & NA    & NA                         \\
                        & PVQ-w \cite{LSVQ}       & 0.827 & 0.828         & 0.711 & 0.739          & 0.791 & 0.795      & 0.770 & 0.807      & NA    & NA                        \\
                        &BVQA-2022 \cite{li2022blindly}    & 0.852 & 0.854         & 0.771 & 0.782          & 0.834 & 0.837      &0.816  & 0.824      & 0.831 & 0.819                     \\    \hdashline
\multirow{4}{*}{\makecell[l]{Existing \\ Transformer \\ Methods}}
                        &StarVQA \cite{xing2022starvqa}      & 0.851 & 0.857         & 0.766 & 0.789          & 0.812  & 0.796     &0.732  & 0.808      & 0.751    & 0.743                     \\
                        &FAST-VQA \cite{wu2022fast}     & 0.876 & 0.877         & 0.779 & 0.814          & 0.842     & 0.844        & 0.765    & 0.782         & 0.794    & 0.784                     \\
                        &DisCoVQA \cite{wu2022discovqa}     & 0.859 & 0.850         & 0.750    & 0.780             & 0.847  & 0.847     & 0.820 & 0.826      & 0.809 & 0.808           \\
                        &DCVQE \cite{li2022dcvqe}       & 0.836 & 0.834         & 0.727 & 0.758          & 0.821 & 0.822      & 0.748 &0.765       & 0.807 & 0.805                   \\
                        \hdashline
\multirow{6}{*}{\makecell[l]{ CLIPVQA \\ Variants}}
                        &CLIPVQA-16-8  & 0.869 & 0.878         & 0.767 & 0.813          & 0.837  &0.844      & 0.811 &0.849       & 0.809 & 0.801        \\
                        &CLIPVQA-16-16 & 0.868 & 0.873         & 0.774 & 0.812          & 0.847  &0.852     & 0.823 & 0.861 & 0.820  & 0.802          \\
                        &CLIPVQA-16-32 & \uwave{0.881} & \uwave{0.883}  & 0.782 & 0.827 &\underline{0.867}& \underline{0.871} &\underline{0.843}  &\underline{0.869} &\underline{0.839} &\underline{0.845}      \\
                        &CLIPVQA-16-64 & \underline{0.879} & \underline{0.882}  & \bf{0.793} &\bf{0.834}   & \bf{0.875}&\uwave{0.879} &\uwave{0.845} & \uwave{0.884} & \bf{0.859} & \uwave{0.861}\\  \cdashline{2-12}
                        &CLIPVQA-14-8  & \bf{0.883} & \bf{0.885}  & \uwave{0.785}&\uwave{0.833}    & \uwave{0.868} & \bf{0.885}      &\bf{0.853} & \bf{0.888} & 0.838 & 0.821  \\
                        &CLIPVQA-14-16 & \uwave{0.881} & \bf{0.885}  & \underline{0.784} & \underline{0.832}   & 0.862 & 0.863    &0.818 &0.859 & \bf{0.859} & \bf{0.871}         \\
\Xhline{1pt} %

\end{tabular}
}
\begin{tablenotes}
 \item[1] \textit{The best, second, and third performances are shown in bold, solid underline and wavy underline, respectively. NA indicates not applicable.}
 \end{tablenotes}
 \vspace{-1.5em}
\end{table*}

\subsection{Performance Evaluation on Individual Datasets}
\label{subsec:exp:ind}

In this subsection, we evaluate the VQA performance of the comparison methods and all CLIPVQA variants on individual video datasets, where the training and testing sets are drawn from the same dataset. The experimental results are reported in Table \ref{individual train}, and the following interesting observations can be made. 
\begin{itemize}
\item[$\bullet$] CNNs-based methods perform better than the classical methods because the deep features extracted by CNNs contain more useful information for the VQA task than the handcrafted features. On the other hand, Transformer-based methods show more promising VQA performance than CNNs-based methods, which is mainly due to their prominent temporal modeling capacity by self-attention mechanism while CNNs-based methods are inferior in this specialty.

\item[$\bullet$] By modeling temporal distortion and extracting content quality information, DisCoVQA is more competitive than DCVQE and StarVQA. LSCT-PHIQ relies on a long short-term convolutional Transformer to capture long-term dependence over temporal units, yielding significant performance improvements on KoNViD-1k and YouTube-UGC. Based on a quality-retained sampling scheme to preserve the spatiotemporal sensitivity, FAST-VQA achieves very good performance on LSVQ and LSVQ-1080p. These improvements imply that modeling the spatiotemporal characteristics from videos is crucial for the VQA problem.

\item[$\bullet$] In terms of the proposed method, CLIPVQA outperforms all the baselines in terms of both SROCC and PLCC metrics on all datasets, achieving state-of-the-art VQA performance. For example, compared with DisCoVQA, CLIPVQA-14-8 shows improvements of $4.0\%$ SROCC and $7.5\%$ PLCC on LIVE-VQC. On one hand, this indicates that leveraging the natural language as supervision can indeed help estimate in-the-wild video quality. On the other hand, by introducing the video frame perception and the content and language aggregation schemes, our CLIPVQA is able to learn more perceptually meaningful spatiotemporal representations for VQA prediction.

\item[$\bullet$] For the high-resolution videos like LSVQ-1080p and YouTube-UGC containing $4$K videos, our CLIPVQA also demonstrates remarkable VQA performance. From Table \ref{individual train}, we can see that CLIPVQA is still able to surpasses the best baseline on LSVQ-1080p (i.e., FAST-VQA) by achieving $1.8\%$ SROCC and $2.5\%$ PLCC improvements.
This shows that CLIPVQA is consistently effective for high-resolution videos that are continuously increasing in real-world applications.
The main reason is that our method leverages the video frame perception Transformer to simultaneously capture the long-range and long-term dependencies of the spatial and temporal information that are directly related to the video quality. This is especially important for high-resolution videos. With the help of other modules, CLIPVQA beats the competitors. 

\item[$\bullet$]
We can also observe that larger CLIP model (i.e., CLIP-L-$14$) for CLIPVQA leads to better assessment performance. This is because, for a larger CLIP network, more fine-grained patches will be generated, making it easier to capture more quality information. In addition,
appropriately increasing the input number of video frames can also improve the VQA performance to some extent. Nevertheless, these two practices will result in higher computational consumption, as seen in Table \ref{Implementation details}.

\end{itemize}

\subsection{Performance Evaluation Across Datasets}
\label{subsec:exp:cross}
A good VQA method is expected to generalize well to unseen
distortion scenarios.
In this subsection, the performances of comparison methods are evaluated across disparate video datasets so as to verify their generalizability and robustness. In other words, the training and testing sets for a VQA method are from different datasets.
Following the settings in \cite{li2022blindly, wu2022discovqa, wu2022fast}, three groups of experiments are conducted across datasets with different scales, where the CLIPVQA-$16$-$32$ variant is selected for comparison due to its nice performance trade-off between the assessment accuracy and computational overhead.
The results are reported in Tables \ref{Cross-Dataset1}, \ref{Cross-Dataset2}, and \ref{Cross-Dataset3}, from which we have the following observations.
\begin{itemize}
\item[$\bullet$] Table \ref{Cross-Dataset1} presents the cross-dataset results of VQA methods on three relatively small datasets. It can be seen that DisCoVQA \cite{wu2022discovqa} and BVQA-2022 \cite{li2022blindly} can generalize well to the unseen datasets. Nevertheless, our proposed CLIPVQA outperforms them in most cases.
In particular, for the generalization between KoNViD-1k (training) and YouTube-UGC (testing), CLIPVQA obtains improvements of $22.8\%$ SROCC and $37.3\%$ PLCC when compared with BVQA-2022.
This reveals that through the deep features or temporal distortion extraction, one can reach good VQA performance. But, the proposed method is able to learn the spatiotemporal relationships more robustly.

\item[$\bullet$] Tables \ref{Cross-Dataset2} and \ref{Cross-Dataset3} present the results on larger-scale datasets LSVQ and KonViD-150k, respectively. It is observed that the generalization capacity of CLIPVQA is more pronounced on these two larger-scale datasets, invariably surpassing the best competitor in most cases by improving up to $6.6\%$ SROCC and $9.9\%$ PLCC. These cross-domain performance gains demonstrate the favorable generalizability of our proposed method, which is mainly due to the effectiveness of the proposed CLIP-based video frame perception scheme.

\end{itemize}

\begin{table*}[t]
\centering
\caption{\small Performance Evaluation Across Small-scale Datasets.}\label{Cross-Dataset1}
\resizebox{1.9\columnwidth}{!}{
\centering
\small
\begin{tabular}{c ccc| ccc| ccc}
\Xhline{1pt}
Training sets/                & \multicolumn{3}{c}{KoNViD-1k}                        & \multicolumn{3}{c}{YouTube-UGC}    & \multicolumn{3}{c}{LIVE-VQC}       \\ \hdashline
\multirow{2}{*}{Testing sets/}  & LIVE-VQC &  & YouTube-UGC                            & KoNViD-1k                        &      & LIVE-VQC         & KoNViD-1k    & & YouTube-UGC         \\   \cline{2-4} \cline{5-7} \cline{8-10}
                                & SROCC      \quad   PLCC   &   & SROCC  \quad  PLCC           & SROCC \quad   PLCC    &      & SROCC \quad   PLCC   & SROCC \quad   PLCC && SROCC \quad  PLCC\\ \hline
VIDEVAL   \cite{VIDEVAL}        & 0.627     \quad 0.654 &  & 0.372 \quad 0.393                & 0.611  \quad 0.620 && 0.542  \quad 0.553               & 0.625  \quad 0.621  &      & 0.302   \quad 0.318 \\
MDTVSFA     \cite{MDVSFA}       & 0.718   \quad 0.760 & & 0.424  \quad 0.460 & 0.649  \quad 0.646  && 0.582  \quad 0.603 & 0.695  \quad 0.711  && 0.354   \quad 0.388 \\
DisCoVQA  \cite{wu2022discovqa} &{\bf{0.782}}  \quad 0.797 & & 0.415 \quad 0.449 &0.686   \quad 0.697  && 0.661  \quad 0.685 & \bf{0.792} \quad \bf{0.785}&& 0.409  \quad 0.432 \\
BVQA-2022  \cite{li2022blindly} & 0.778  \quad 0.780 &  & 0.475  \quad 0.486 & {\bf{0.783}}  \quad 0.773  && 0.652  \quad 0.675 & 0.738 \quad 0.721&& \bf{0.596}  \quad \bf{0.596} \\
CLIPVQA    & 0.771 \quad \bf{0.827}& &\bf{0.583} \quad \bf{0.667}   &0.722 \quad \bf{0.778} & & \bf{0.663} \quad 0.698     & 0.703 \quad 0.728  & & 0.488\quad 0.533 \\ \Xhline{1pt}
\end{tabular}
}
\vspace{-1.5em}
\end{table*}

\begin{table}[t]
\centering
\caption{\small Performance Evaluation Across Medium-scale Datasets.}\label{Cross-Dataset2}
\resizebox{1\columnwidth}{!}{
\large
\begin{tabular}{l:lcccc}

\Xhline{1pt} %
\multicolumn{2}{c}{Training sets/ }    &\multicolumn{4}{c}{LSVQ}    \\
\hdashline%
\multicolumn{2}{c}{Testing sets/ }       & \multicolumn{2}{c}{KoNViD-1k  }             &  \multicolumn{2}{c}{ LIVE-VQC  }   \\  \hline
   Groups & Methods                      &  SROCC & PLCC  &  SROCC & PLCC           \\           \hline
\multirow{3}{*}{Classical Methods}
                        & BRISQUE \cite{mittal2012no}       & 0.646 & 0.647         & 0.524 & 0.536     \\
                        & TLVQM \cite{TLVQM}         & 0.732 & 0.724         & 0.670 & 0.691       \\
                        & VIDEVAL \cite{VIDEVAL}        & 0.751 & 0.741         & 0.630 & 0.640                \\  \hdashline
\multirow{4}{*}{CNNs-based Methods}
                        & VSFA  \cite{VSFA}         & 0.784 & 0.794         & 0.734 & 0.772                \\
                        & PVQ-wo  \cite{LSVQ}        & 0.781 & 0.781         & 0.747 & 0.776                \\
                        & PVQ-w   \cite{LSVQ}        & 0.791 & 0.795         & 0.770 & 0.807               \\
                        & BVQA-2022 \cite{li2022blindly}     & 0.834  & 0.837 & 0.816 & 0.824               \\   \hdashline

\multirow{3}{*}{Transformer Methods}
                        &FAST-VQA \cite{wu2022fast}       & 0.859 & 0.855  & \bf{0.823} & 0.844                \\
                        &DisCoVQA \cite{wu2022discovqa}       & 0.846  & 0.849     & \bf{0.823}  & 0.837                \\
                        &CLIPVQA  & \bf{0.864} &\bf{0.887}     & 0.781 & \bf{0.871}                \\
\Xhline{1pt} %
\end{tabular}
}
\vspace{-1em}
\end{table}

\begin{table*}[t]
\centering
\caption{\small Performance Evaluation Across Large-scale Datasets.}\label{Cross-Dataset3}
\begin{tabular}{l:lcc|cc|cc}
\Xhline{1pt} %
\multicolumn{2}{c}{Training sets/ } &\multicolumn{4}{c}{KonViD-150k-A}       \\
\hdashline%
\multicolumn{2}{c}{Testing sets/ }              & \multicolumn{2}{c}{LIVE-VQC}   & \multicolumn{2}{c}{KoNViD-1k}   & \multicolumn{2}{c}{KonViD-150k-B}        \\  \hline
   Groups & Methods                             &  SROCC & PLCC       &  SROCC & PLCC      &  SROCC & PLCC                         \\           \hline
\multirow{2}{*}{CNNs-based Methods}
                        & VSFA  \cite{VSFA}     & 0.708  & 0.733             & 0.801  & 0.815                    & 0.813  & 0.808                         \\
                        & MLSP-FF \cite{gotz2021konvid}   & 0.738&0.754      & 0.828  & 0.821      & 0.827  & 0.852                        \\   \hdashline
\multirow{2}{*}{Transformer Methods}
                        &DisCoVQA \cite{wu2022discovqa} & 0.751&0.766  & \bf{0.843}   &\bf{0.841}        & 0.845  & 0.858                            \\
                        &CLIPVQA         & \bf{0.801}&\bf{0.842}    & 0.810    &0.821         & \bf{0.866}  & \bf{0.885}                     \\
\Xhline{1pt} %
\end{tabular}
\end{table*}


\subsection{Performance Evaluation with Fine-tuning}
\label{subsec:exp:fine}
At present, an absolute majority of VQA datasets is very small. Learning-based VQA methods usually suffer from the lack of large-scale datasets for network training. To alleviate this issue, pre-training followed by fine-tuning has been a widely studied learning paradigm. In this part, the performances of comparison methods are evaluated in a pre-training-fine-tuning manner, where the CLIPVQA-$16$-$32$ variant is employed. As far as we know, LSVQ is the second-largest VQA dataset to date. Meanwhile, for a fair comparison, we adopt the same LSVQ dataset as used in the top-performing FAST-VQA method \cite{wu2022fast} for pre-training and five small-scale datasets representing various scenarios for fine-tuning. Specifically, the small-scale datasets include LIVE-VQC (real-world mobile photography with resolutions in $240$P-$1080$P), KoNViD-1k (having various contents collected online with all in $540$P), CVD2014 (containing synthetic in-capture distortions with all in $720$p), LIVE-Qualcomm (having selected types of distortions with all in $1080$P) and YouTube-UGC (owning user-generated contents, including the computer graphic contents, with all in $360$P-$2160$P). The performance results are reported in Table \ref{finetune train}, which leads to the following observations.

\begin{itemize}
\item[$\bullet$] Consistent with previous observations, the proposed CLIPVQA yields superior performance compared with the state-of-the-art methods on almost all datasets except CVD2014, which we believe mainly benefits from the proposed CLIP-based attention paradigm. FAST-VQA employs a fragment-based sampling scheme to preserve the video quality, which can well deal with the synthetic distortions to some extent, thus having good performance on CVD2014.

\item[$\bullet$] By comparing Table \ref{individual train} with Table \ref{finetune train}, we can find that after fine-tuning, the VQA performances of all comparison methods on all datasets are further improved, even on the YouTube-UGC dataset containing a significant amount of $4$K (i.e., $2160$P) videos. This implies that pre-trained CLIPVQA can serve as a powerful backbone to facilitate the VQA-related downstream tasks.
\end{itemize}

\begin{table*}
\centering
\caption{\small Performance Results with Fine-tuning.}\label{finetune train}
\resizebox{1.9\columnwidth}{!}{
\small
\begin{tabular}{l:lcccccccccc}
\Xhline{1pt} %
\multicolumn{2}{c}{Testing sets/ }              & \multicolumn{2}{c}{LIVE-VQC}   & \multicolumn{2}{c}{KoNViD-1k }        & \multicolumn{2}{c}{CVD2014  } &  \multicolumn{2}{c}{ LIVE-Qualcomm  }   &  \multicolumn{2}{c}{ YouTube-UGC  }\\  \hdashline
   Groups & Methods                             &  SROCC & PLCC       &  SROCC & PLCC       &  SROCC & PLCC        &  SROCC & PLCC    &  SROCC & PLCC       \\           \hline
\multirow{3}{*}{\makecell[l]{ Classical \\ Methods}}
                        & TLVQM \cite{TLVQM}                  & 0.799 & 0.803        & 0.773 & 0.768           & 0.830 & 0.850         & 0.770 & 0.810    & 0.669 & 0.659    \\
                        & VIDEVAL \cite{VIDEVAL}              & 0.752 & 0.751        & 0.783 & 0.780           & NA & NA               & NA & NA          & 0.779 & 0.773   \\
                        & RAPIQUE  \cite{tu2021rapique}       & 0.755 & 0.786        & 0.803 & 0.817           & NA & NA               & NA & NA          & 0.759 & 0.768          \\  \hdashline

\multirow{6}{*}{\makecell[l]{ CNNs-based \\ Methods}}
                        & VSFA \cite{VSFA}                 & 0.773 & 0.795        & 0.773 & 0.775           & 0.870 & 0.868         & 0.737 & 0.732    & 0.724 & 0.743            \\
                        & PVQ \cite{LSVQ}                  & 0.827 & 0.837        & 0.791 & 0.786           & NA & NA               & NA & NA          & NA & NA           \\
                        & GST-VQA \cite{chen2021learning}  &NA &NA                &0.814 &0.825             &0.831 &0.844           &0.801 &0.825      &NA &NA         \\
                        & CoINVQ   \cite{wang2021rich}     &NA &NA                &0.767  &0.764             &0.831 &0.844          &NA &NA            &0.816  &0.802   \\
                        &CNN+TLVQM \cite{korhonen2020blind}  &0.825 &0.834          &0.816 &0.818             &0.863 &0.880           &0.810 &0.833      &NA  &NA   \\
                        &BVQA-2022  \cite{li2022blindly}     &0.831 &0.842          &0.834 &0.836             &0.872 &0.869           &0.817 &0.828      &0.831 &0.819    \\ \hdashline
\multirow{4}{*}{\makecell[l]{Transformer \\ Methods}}
                        &Full-res Swin-T \cite{wu2022fast}       &0.799 &0.808          &0.841 &0.838             &0.868 &0.870           &0.788 &0.803      &0.798 &0.796   \\
                        &FAST-VQA-M \cite{wu2022fast}            &0.803 &0.828          &0.873 &0.872             &0.877 &0.892           &0.804 &0.838      &0.768 &0.765   \\
                        &FAST-VQA \cite{wu2022fast}              &0.849&0.865&0.891     &0.892 &\bf{0.891} &\bf{0.903}      &0.819 &0.851      &0.855 &0.852   \\
                        &CLIPVQA          &\bf{0.870}&\bf{0.892}&\bf{0.907} & \bf{0.912} &0.883 &0.888        & \bf{0.833}  & \bf{0.872}         & \bf{0.881}  &\bf{0.883}   \\
\Xhline{1pt} %
\end{tabular}
}
\end{table*}

\subsection{Qualitative Analyses}
\label{subsec:exp:mos}
\begin{figure}[t]
\centering
\includegraphics[width=0.45\textwidth]{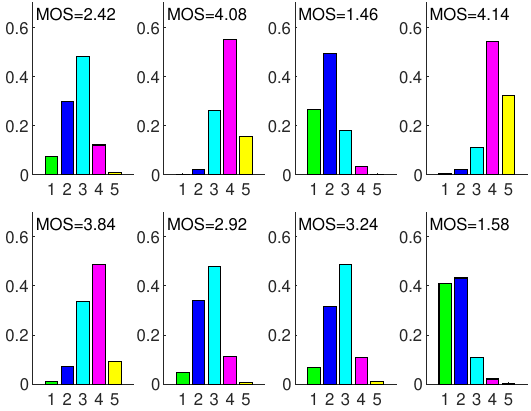} 
\caption{\small A visualization of the predicted distributions by CLIPVQA on a set of randomly selected samples from KoNViD-1k dataset.}\label{pro}
\vspace{-1em}
\end{figure}

In this subsection, we provide qualitative analyses for CLIPVQA by visualizing its predicted distributions on a set of randomly selected samples from KoNViD-1k dataset, as shown in Fig.~\ref{pro}. 
The x-axis of histograms represents the quality scales and the y-axis denotes the predicted probability for each scale. We can observe that the largest value of the predicted distributions by our method highly matches the MOS value, which suggests that the proposed CLIPVQA is a very good quality estimator for the VQA task, confirming its prominent performance in SROCC and PLCC metrics.


\begin{table*}[htp]
\centering

\caption{\small Performance Results of CLIPVQA with and without Fusion Tokens and SAT Module. }\label{ablation1}
\begin{tabular}{l|lc|cc|cc|cc}
\Xhline{1pt}
\multirow{2}{*}{\makecell[l]{Testing Set/ \\ Methods/ Metric/}} &\multicolumn{2}{c}{KoNViD-1k}&\multicolumn{2}{c}{LIVE-VQC}&\multicolumn{2}{c}{YouTube-UGC } &\multicolumn{2}{c}{LSVQ }\\ \cline{2-9}
                                                             & SROCC      &PLCC                 & SROCC   &PLCC            & SROCC &PLCC        & SROCC &PLCC          \\  \hdashline
Without Fusion Tokens     & 0.833      & 0.845               & 0.788   &0.853           & 0.831 &0.822       & 0.864 &0.872       \\  \hdashline
Without SAT Module   & 0.845      & 0.858               & 0.834   & \bf{0.871}           & 0.835 &0.833       & 0.878 &0.880       \\   \hdashline
Without Fusion Tokens and SAT module & 0.822      & 0.830               & 0.755   &0.825           & 0.825 &0.813       & 0.851 &0.857        \\  \hdashline
Full version of CLIPVQA  & \bf{0.867}      &\bf{ 0.871}               & \bf{0.843}   & 0.869           & \bf{0.839} & \bf{0.845}       & \bf{0.881} & \bf{0.883}      \\
\Xhline{1pt} %
\end{tabular}

\end{table*}

\begin{table*}[t]
\centering
\caption{\small Performance Results of CLIPVQA with and without Video Quality Language Supervision.}\label{ablation2}
\begin{tabular}{l|cc|cc|cc|cc}
\Xhline{1pt}
\multirow{2}{*}{\makecell[l]{Testing Set/ \\ Methods/ Metric/}} &\multicolumn{2}{c}{KoNViD-1k} &\multicolumn{2}{c}{LIVE-VQC} &\multicolumn{2}{c}{YouTube-UGC } &\multicolumn{2}{c}{LSVQ } \\ \cline{2-9}
                                           & SROCC &PLCC      & SROCC &PLCC       & SROCC &PLCC      & SROCC &PLCC                   \\  \hdashline
 Without Quality Language                  & 0.844 & 0.851    & 0.782 & 0.821     & 0.811 & 0.801    & 0.870 & 0.872                 \\ \hdashline
 Long Quality Language                     & 0.860 & \bf{0.873}    & 0.755 & 0.836     & 0.829 & 0.836    & 0.873 & 0.876             \\ \hdashline
 Short Quality Language                    & \bf{0.867} & 0.871    & \bf{0.843}   & \bf{0.869}    & \bf{0.839} & \bf{0.845}     & \bf{0.881} & \bf{0.883}    \\ \hdashline
Without Converting MOS into Language & 0.838 & 0.851   & 0.791   & 0.842     & 0.801   & 0.812& 0.835   & 0.854     \\
\Xhline{1pt} 
\end{tabular}
\end{table*}

\begin{table*}[t]
\centering
\caption{\small Performance Results of CLIPVQA using Different Frame Token Representation Strategies for VAT Module and the Case without using VAT Module.}\label{ablation3}
\begin{adjustbox}{max width=\textwidth}
\begin{tabular}{l|cc|cc|cc|cc}
\Xhline{1pt}
\multirow{2}{*}{\makecell[l]{Testing Set/ \\ Methods/ Metric/}} &\multicolumn{2}{c}{KoNViD-1k} &\multicolumn{2}{c}{LIVE-VQC} &\multicolumn{2}{c}{YouTube-UGC } &\multicolumn{2}{c}{LSVQ } \\ \cline{2-9}
                                           & SROCC &PLCC      & SROCC &PLCC       & SROCC &PLCC      & SROCC &PLCC               \\  \hdashline
$\mathbf{MEAN}$ Strategy    & 0.853 & \bf{0.878}    & 0.821 & 0.868     & 0.830 & 0.836    & 0.876 & 0.878            \\ \hdashline
Frame-by-frame-fusion Strategy  & 0.866 & 0.870    & 0.833 & 0.867     & \bf{0.844} & \bf{0.852}    & 0.877 & 0.879           \\ \hdashline
$\mathbf{CONV2d}$ Strategy   & \bf{0.867} & 0.871    & \bf{0.843}   & \bf{0.869}    & 0.839 &0.845     & \bf{0.881} & \bf{0.883}        \\ \hdashline
Without VAT Module         & 0.860   & 0.869     & 0.821   & 0.855   & 0.757  & 0.809  & 0.870 & 0.871    \\
\Xhline{1pt} %
\end{tabular}
\end{adjustbox}
\end{table*}

\begin{table*}[t]
\centering
\caption{\small Performance Results of CLIPVQA with Different Loss Functions.}\label{a5}
\begin{adjustbox}{max width=\textwidth}
\begin{tabular}{l|cc|cc|cc|cc}
\Xhline{1pt}
\multirow{2}{*}{\makecell[l]{Testing Set/ \\ Methods/ Metric/}} &\multicolumn{2}{c}{KoNViD-1k} &\multicolumn{2}{c}{LIVE-VQC} &\multicolumn{2}{c}{LSVQ}&\multicolumn{2}{c}{YouTube-UGC} \\ \cline{2-9}
                                           & SROCC & PLCC & SROCC & PLCC & SROCC& PLCC & SROCC & PLCC  \\  \hdashline
 VR Loss                   & \textbf{0.867} & \textbf{0.871}  & \textbf{0.843}& \textbf{0.869}   & \textbf{0.881}& \textbf{0.883}  & \textbf{0.839}& \textbf{0.845}        \\ \hdashline   
 CE Loss                    & 0.769 & 0.598           & 0.714   & 0.609          & 0.808  & 0.796        & 0.777   & 0.769            \\
\Xhline{1pt} %
\end{tabular}
\end{adjustbox}
\end{table*}


\begin{table*}[htp]
\centering
\caption{\small Computational Consumption of Compared VQA Methods.}\label{train time}
\begin{tabular}{c|l|lc|cc|cc|cc}
\Xhline{1pt}
\multirow{2}{*}{ GPU} &\multirow{2}{*}{Methods} &\multicolumn{2}{c}{540p}    &\multicolumn{2}{c}{720p}  &\multicolumn{2}{c}{1080p}  &\multicolumn{2}{c}{2160p} \\ \cline{3-10}
                    &  & FLOPs(G)      &Time(s)                 &FLOPs(G)       &Time(s)                  &FLOPs(G)       &Time(s)  &FLOPs(G)       &Time(s) \\ \cline{1-10}
\multirow{4}{*}{\makecell[l]{Tesla V100}}
 & VSFA \cite{VSFA}              & 10249         & 2.603                  & 18184         &3.571                    & 40919         &11.14  & NA         &NA \\
 & PVQ \cite{LSVQ}                & 14646         & 3.091                  & 22029         &4.143                    & 58501         &13.79  & NA         &NA  \\
 & BVQA-2022 \cite{li2022blindly}          & 28176         & 5.392                  & 50184         &10.83                    & 112537        &27.64  & NA         &NA   \\
 & FAST-VQA \cite{wu2022fast}          & 279           &0.044                   &279            &0.043                    & 279           &0.045  & NA         &NA  \\  \hdashline
\multirow{3}{*}{\makecell[l]{RTX 3090}}
 & StarVQA \cite{xing2022starvqa}          & 590           &0.212                   &590            &0.483                    & 590           &0.884  & 590         &1.781  \\
 & FAST-VQA \cite{wu2022fast}          & 279           &0.124                   &279            &0.294                    & 279           &0.545  & 279         &1.033  \\
 & CLIPVQA      & 574           &0.202                   &574                &0.463                    & 574           &0.874  & 574         &1.623  \\  \Xhline{1pt}
\end{tabular}
\begin{tablenotes}
 \item[1] \textit{NA indicates not applicable.}
 \end{tablenotes}
\end{table*}

\subsection{Ablation Studies}
\label{subsec:exp:ablation}

In this subsection, we conduct ablation studies to investigate the contribution of each module of the proposed CLIPVQA to the final VQA performance. 
We adopt CLIPVQA-$16$-$32$ as a full version of the proposed method in experiments.

\textbf{Fusion Tokens and SAT Module} 
We first study the effectiveness of the proposed fusion token mechanism and spatiotemporal quality aggregation Transformer (SAT). Note that fusion tokens contain spatiotemporal quality-related information, and the SAT module is used to generate video-level representation.
To this end, we consider three variants of CLIPVQA: 1) The first one is without the fusion tokens, leaving the rest of the network unchanged. 2) The second one removes the SAT module, where the other parts remain unchanged. 3) For the third variant, fusion tokens and SAT modules are removed from CLIPVQA, with other components remaining unchanged.
All variants are trained with the vectorized regression loss function (i.e., Eq.~\ref{loss}).
The performance results of each variant are presented in Table \ref{ablation1}.

Based on the results, we can observe that removing fusion tokens leads to a marked decrease in performance on all datasets. This is because the fusion tokens encode both the intra-frame spatial and inter-frame temporal information, comprising rich and diverse quality features that are closely relevant to the whole video quality. Thus, removing them results in the accuracy degradation.
On the other hand, the lack of SAT module also causes a drop in performance on almost all datasets.
The reason is that the SAT module is used to aggregate the pseudo-MOS and fusion tokens for producing the video-level quality representation.
Without it, CLIPVQA cannot integrate this diverse quality-related information to better serve the VQA task.
Note that the performance decline is less pronounced on YouTube-UGC, which is mainly due to the less in-capture temporal distortions.
Obviously, removing both the fusion tokens and SAT module from CLIPVQA significantly degrades the VQA performance on all datasets. The performances decrease on average by $6.0\%$ SROCC and $4.5\%$ PLCC.
These demonstrate the importance of the fusion tokens and SAT module in CLIPVQA, which are indeed effective for VQA problem.

\textbf{Video Quality Language Supervision} 
We then investigate the role of the video quality language supervising scheme in VQA scenario. To achieve this, we conduct experiments by constructing four variants for our CLIPVQA method. The first one is to directly use MOS for encoding without converting the MOS into quality language while other components are kept unchanged. 
The second one is without using any quality language for supervision.
Since the language supervision module (i.e., MOS2Language encoder) and VAT module need to be used together, thus both MOS2Language encoder and VAT module are removed for this variant.
The third and fourth variants utilize long and short video-quality natural language descriptions (i.e., Tables \ref{text description-1} and \ref{text description-2}) for embedding, respectively. 
The performance results are reported in Table \ref{ablation2}.
We can easily see that the first two CLIPVQA variants perform worse than the last two variants on all datasets. The one without converting MOS into language performs the worst, while the variant using short-quality language descriptions as supervision is the best.

The above results indicate that using quality language descriptions as supervision along with the VAT module for generating video-language representation well serves the purpose of enhancing the VQA performance. Directly using MOS for encoding cannot achieve this goal, since the information contained in numerical values is quite limited. 
Thus, the MOS2Language encoder is equally
important as other components to the final performance of framework.
The short description has advantages over the long one, which is probably because the short description makes the network concentrate more on the most important information.

\textbf{Frame Token Representation and VAT Module}
Next, we conduct experiments to verify the impacts of different frame token representations for the VAT module as well as the effectiveness of the whole VAT module.
In the full version of CLIPVQA, we first perform a two-dimensional convolution operation for the $N$ frame tokens and then feed the compressed representation to the VAT module for further processing. In this ablation study, we design two more variants for comparison: 1) One is averaging the $N$ frame tokens and then using the averaged result as input for the VAT module without any convolutional operations, called $\mathbf{MEAN}$ strategy. 2) The other is to directly feed the tokens of each frame to the VAT module and then fuse the outputs of all the frames from VAT to achieve the final video-language representation. Furthermore, to validate the significance of the entire VAT module, we design the last variant that directly uses the quality language encoding $\mathcal{Y}_{t}$ to replace the video-language representation $\mathcal{\tilde Y}_{v}$ in Eq.18, i.e., without utilizing the VAT module.

The performance result of each variant is shown in Table \ref{ablation3}.
We observe that without using the VAT module, it performs the worst in terms of the two metrics on almost all datasets, which suggests that only using quality language encoding as supervision is not enough for the VQA task. 
Quality language encoding needs to be aggregated with the video content information so as to better serve the final VQA prediction, which thereby confirms the effectiveness of the proposed VAT module for the CLIPVQA framework.
This design also distinguishes our CLIPVQA from the other CLIP-based methods like CLIP-IQA \cite{wang2023exploring} and LIQE \cite{zhang2023blind}. For the variants with VAT module, the performance yielded by utilizing the $\mathbf{MEAN}$ strategy is not competitive when compared with the full version employing the $\mathbf{CONV2d}$ action. Although the $\mathbf{MEAN}$ strategy reduces computational consumption, it impairs the performance of the model. If we use the frame-by-frame strategy followed by a fusion operation, the performance is encouraging on some datasets, however its computational consumption is very large. This validates our choice of $\mathbf{CONV2d}$ for the frame token strategy.

\textbf{Vectorized Regression Loss Function}
We finally study the efficacy of the vectorized regression (VR) loss function used in CLIPVQA. To realize this, we execute an experiment by comparing the VR loss function with the cross-entropy (CE) loss function (i.e., the default loss strategy in the original CLIP). The performance results of CLIPVQA variants with these two loss functions are reported in Table~\ref{a5}. We can observe that the CLIPVQA with VR loss outperforms the one with CE loss by a large margin on all datasets, achieving improvements of
up to $18.1\%$ SROCC and $45.7\%$ PLCC, respectively. This indicates that the VR loss function is highly suitable for our proposed framework for effective end-to-end optimization, enabling CLIPVQA to obtain exceptional performance on VQA tasks.


\subsection{Computational Complexity}
\label{subsec:exp:comp}

In this subsection, we perform experiments to test the computational efficiency of CLIPVQA.
We compare the FLOPs and actual running times on GPUs (average of one hundred runs per sample) of CLIPVQA-$16$-$32$ with some state-of-the-art CNNs-based VQA and Transformer VQA approaches on the videos with different resolutions.
The results including two parts are presented in Table \ref{train time}. One part is tested on a GPU of Tesla V100, which is for the competitors. The other part is executed on a GPU of RTX 3090, which is for StarVQA, FAST-VQA and CLIPVQA.
From the results, we can see that FAST-VQA requires few computing resources while our method also exhibits very comparable efficiency especially on high-resolution videos. The computational consumption of StarVQA is similar to that of our CLIPVQA, but its performance is not competitive.
This further manifests that the proposed CLIPVQA has a superior performance trade-off between efficiency and effectiveness compared with other VQA methods.

\section{Conclusion}
\label{sec:conclusion}
This paper has presented an efficient and effective CLIP-based Transformer method for VQA scenario, where CLIP is a contrastive language-image pre-training scheme. First, the proposed method extracts the frame-level spatiotemporal quality features by using a frame perception Transformer. Then, the video-level representation is obtained utilizing a spatiotemporal quality aggregation network.
To make use of the natural language information of video quality, we encode the language descriptions of videos through a CLIP-based encoder and then achieve the video-language representation for videos through a video content and language aggregation module.
The final prediction vector is obtained by fusing both the video-level and video-language representations via a feature fusion operation.
Finally, a vectorized regression loss function is employed for the entire network optimization.
Comprehensive experiments on diverse video datasets are conducted to evaluate the performance of the proposed method. The results demonstrate the state-of-the-art VQA performance, favorable generalizability and robustness of the proposed approach.
This paper broadens the applications of CLIP to the VQA scenario and shows its feasibility and effectiveness, which actually points out an exciting and promising research direction for the objective video quality assessment community.

\bibliographystyle{IEEEtran} 
\bibliography{bare_jrnl_compsoc}

\end{document}